\documentclass{article}

\PassOptionsToPackage{numbers, compress}{natbib}

\usepackage[preprint]{neurips_2026}

\usepackage[utf8]{inputenc} 
\usepackage[T1]{fontenc}  
\usepackage{hyperref}    
\usepackage{url}      
\usepackage{booktabs}    
\usepackage{amsfonts}    
\usepackage{longtable}
\usepackage{nicefrac}    
\usepackage{wrapfig}
\usepackage{microtype}   
\usepackage{xcolor}     
\usepackage{graphicx}
\usepackage{algorithm}
\usepackage{algpseudocode}
\usepackage{amsmath}
\usepackage{amsthm}
\usepackage{array}
\usepackage{multirow}
\usepackage{subcaption}
\newtheorem{theorem}{Theorem}

\newtheorem{corollary}{Corollary}
 
\title{DARTS: Targeting Prognostic Covariates in Budget-Constrained Sequential Experiments}

\author{Kateryna Husar
 \\
 Department of Statistical Science\\
 Duke University\\
 Durham, NC 27705 \\
 \And
Alexander Volfovsky \\
 Department of Statistical Science\\
 Duke University\\
 Durham, NC 27705}

\begin{document}

\maketitle

\begin{abstract}
Randomized controlled trials typically assume that prognostic covariates are known and available at no cost. In practice, obtaining high-dimensional pretreatment data is costly, forcing a trade-off between covariate-adaptive precision and a measurement budget. We introduce Dynamic Adaptive Rerandomization via Thompson Sampling (DARTS), which treats covariate acquisition as a sequential optimization problem embedded within a design-based causal inference task.
A budgeted combinatorial Thompson sampler learns which covariates are most prognostic across successive batches; selected covariates then drive rerandomization and regression adjustment to reduce batch-level average treatment effect variance. Our primary theoretical contribution is a decoupling result: adaptive covariate selection based on past batches preserves batch-level randomization validity, and the cumulative inverse-variance weighted estimator achieves at least nominal asymptotic coverage. We further derive a Bayes risk bound for the acquisition layer that matches the minimax lower bound up to logarithmic factors. Empirically, DARTS systematically concentrates the budget on informative features, significantly closing the efficiency gap to oracle designs while maintaining strict inferential validity.
\end{abstract}

\section{Introduction}
A common challenge in experimental design is the ``Individual Trial Fallacy'': while complete randomization ensures unbiasedness over the design space, any single realized allocation may exhibit significant conditional bias due to covariate imbalance, particularly in small samples or high-dimensional settings. 
To mitigate this, researchers utilize pretreatment data to enforce balance during the design phase (e.g., via rerandomization or stratification) or to adjust for imbalances during analysis. 
But the assumption that high-quality pretreatment data are readily available is often unrealistic. 
In practice, acquiring such data is rarely cost-free, constrained by medical diagnostic expenses, survey fatigue, or strict computational overhead \citep{turney2002types,li2021active, janisch2019classification}. 
Furthermore, the efficacy of covariate-adaptive methods diminishes as the number of prognostic factors grows. 
When the covariate space is high-dimensional relative to the sample size, achieving simultaneous balance across all dimensions becomes mathematically improbable and computationally expensive.
If the most prognostic factors are known a priori, existing literature provides robust frameworks for their inclusion \citep{morgan2012rerandomization, li2020rerandomization, zhou2018sequential, bugni2018inference, yang2024sequential, yang2023balancing, zhang2021response, li2026design, molitor2025anytime}. 
Yet, in many settings, these factors are unknown. 
In such cases, we must learn which variables to prioritize while adhering to a strict measurement budget. 
Instead of examining all of the data at once, it is natural to consider sequential batches and incorporate learning steps. 
Moreover, we often already encounter settings where data arrive in groups: participants in studies might continuously enroll over months or years, or interventions are implemented district-by-district. 
By treating the experiment as a series of batches, we can iteratively update our understanding of the covariate-outcome relationship. 

A substantial literature addresses related problems across three communities. 
We propose the Dynamic Adaptive Rerandomization via Thompson Sampling (DARTS) framework, which treats covariate acquisition as a sequential optimization problem embedded within a valid randomized trial design, and organize the relevant work to clarify where DARTS sits relative to each.

\textbf{Covariate-adaptive randomization and design-based inference.} The foundational approach of \citet{morgan2012rerandomization} and subsequent work (\citet{bugni2018inference,yang2023balancing, yang2024sequential,zhang2021response,zhou2018sequential}) address covariate balance under a \emph{known} prognostic set, either by filtering allocations via Mahalanobis distance or by  tilting assignment probabilities as units arrive; \citet{li2026design} and \citet{molitor2025anytime} additionally incorporate anytime-valid inference. DARTS is complementary: it answers the prior question of \emph{which} covariates to balance on when they are unknown and costly to measure.

\textbf{Bandits with knapsacks and combinatorial semi-bandits.} The budgeted combinatorial bandit framework of \citet{badanidiyuru2018bandits}, extended by \citet{das2022budgeted} to the semi-bandit setting with LP relaxations, provides the algorithmic backbone for DARTS's acquisition layer. \citet{liu2023variable} introduce Thompson sampling for variable selection using Beta posteriors; DARTS extends this to a budgeted combinatorial setting with an explicit causal inference objective. The key question is not how to run a  budgeted semi-bandit efficiently, but whether doing so inside a sequential RCT 
preserves causal validity.

\textbf{Adaptive feature acquisition and active experimental design.} A growing literature treats feature acquisition as a sequential decision problem  (\citet{janisch2019classification,li2021active,shim2018joint,covert2023learning}), learning policies that minimize  prediction error under feature costs. 
These methods target predictive accuracy for individual units and carry no randomization validity requirement: the acquisition policy may freely use outcome data. 
DARTS maintains a strict separation between acquisition policy and outcomes.

\textbf{Contributions.} DARTS's contributions are ordered as follows:

\textbf{First and foremost}, we introduce a new problem formulation: adaptive covariate acquisition as part of sequential RCT design under a measurement budget. 
\textbf{Second}, we establish a design-based validity guarantee: adaptive covariate selection based on past batches does not compromise batch-level randomization validity (Theorem~\ref{thm:unbiased}), and the cumulative inverse-variance weighted estimator achieves at least nominal asymptotic coverage (Theorem~\ref{thm:conserv}).  
\textbf{Third},  we provide supporting bandit theory for the acquisition layer: a Bayes risk bound and a minimax lower bound for the binary-reward case, whose rates match up to a logarithmic factor, and an extended bound for the fractional reward scheme used in practice (Theorems \ref{thm:regret} and \ref{thm:lower_bound}).
\textbf{Fourth}, we validate empirically that adaptive covariate acquisition meaningfully closes the gap to oracle rerandomization across a range of DGPs and budget regimes, and that the Thompson sampler learns to concentrate on the true prognostic covariates. 

\section{Sequential Randomized Trial}
\label{sec:srt}
\paragraph{Randomized control trial.}
We consider an experiment over $T$ batches of $n$ units each. Let $Z_i(t) \in \{0,1\} $ denote the treatment assignment for unit $i $ in batch $t $, with $N_1(t) = N_0(t) = n/2 $ being the number of treated and control units respectively.

For each unit, we define a pair of potential outcomes: $Y_i(1,t) $, the outcome that would be observed under treatment, and $Y_i(0,t) $, the outcome that would be observed under control. 
We impose Stable Unit Treatment Value Assumption (SUTVA): potential outcomes of unit $i$ depend only on its own treatment, and treatment is well-defined and uniform across units. The observed outcome for unit $i $ defined by $Y_i(t) = Z_i(t) Y_i(1,t) + (1-Z_i(t))Y_i(0,t) $, where the only source of randomness comes from treatment assignment.

Our target estimand is the Average Treatment Effect (ATE), $\tau = \mathbb{E}[Y_i(1,t) - Y_i(0,t)] $, common across all batches. Since both potential outcomes are never jointly observed, a natural within-batch estimator is the difference-in-means: 
\[
\hat\tau_t^{DiM} = \tfrac{1}{N_1(t)}\sum_{i=1}^n Z_i(t)Y_i(t) - \tfrac{1}{N_0(t)}\sum_{i=1}^n(1-Z_i(t))Y_i(t).
\]
This is trivially unbiased, but  can be imprecise when an unlucky allocation produces covariate imbalance between arms.

\paragraph{Rerandomization, covariate adjustment, and precision gains.}
To improve precision within each batch, we can leverage pre-treatment covariates $X_i(t)\in\mathbb{R}^p$. In the \emph{design phase}, we can rerandomize using a generalization of \citep{morgan2012rerandomization}: 
for batch $t$, we draw $\mathcal{K}$ candidate allocations and select
\[
z^* = \arg\min_z\; d_M(z, \mathbf{X}(t)) 
= (\bar X_1(t)-\bar X_0(t))^\top\hat\Sigma(t)^{-1}(\bar X_1(t)-\bar X_0(t)),
\]
where $\bar X_1(t)$, $\bar X_0(t)$ are the treatment and control covariate means and $\hat\Sigma(t)$ is the pooled sample covariance within batch $t$. 
We accept randomization $Z(t)$ for batch $t$ by tossing a fair coin between $z^*$ and $\mathbf{1}-z^*$ to preserve exact marginal balance.
In the \emph{analysis phase}, Lin-adjusted OLS on $Y(t)\sim Z(t)\times X(t)$ yields $\hat\tau_t$ with HC2 variance estimate $\hat v_t$. 

Combining rerandomization and Lin adjustment never hurts precision under a mirror-symmetry condition \citep{li2020rerandomization} which our minimization-based design satisfies.

\paragraph{Combining batch estimates.}
Across batches, we combine estimates using precision-weighted meta-analysis.
After observing batch $t $, the cumulative ATE estimate is updated as:
\begin{equation}
\label{eq:meta}
\hat{\sigma}_t^2 = \left(\hat{\sigma}_{t-1}^{-2} + 
  \hat{v}_t^{-1}\right)^{-1}, \qquad 
\hat{\mu}_t = \hat{\sigma}_t^2\left(\tfrac{\hat{\mu}_{t-1}}
  {\hat{\sigma}_{t-1}^2} + \tfrac{\hat{\tau}_t}{\hat{v}_t}
  \right), \qquad \hat\mu_0 = 0,
\end{equation}
which is the standard inverse-variance weighted estimator. This sequential update is equivalent to pooled OLS on all observed data when the batch-level variance estimates are well-calibrated. Each batch contributes to the cumulative estimate in proportion to its precision, so batches with better-balanced covariates and hence smaller $\hat{v}_t $ naturally receive greater weight.

\begin{theorem}[Unbiasedness of sequential ATE estimator]
\label{thm:unbiased}
Let $\mathcal F_{t-1} $ denote the sigma-field generated by all covariate measurements, treatment assignments, and outcomes through batch $t-1 $. Suppose the selected covariate set $S_t $ is chosen by an $\mathcal F_{t-1} $-measurable rule. Assume SUTVA and that the units in batch $t $ are drawn i.i.d. from the target population independently of $\mathcal F_{t-1} $. 
Let 
\[
P_t(Z(t)=z\mid \mathbf X(t),\mathcal F_{t-1})
= P_t(Z(t)=\mathbf 1-z\mid \mathbf X(t),\mathcal F_{t-1})
\]
for every feasible assignment vector $z $. Then the unadjusted
difference-in-means estimator satisfies
$\mathbb E\!\left[
\hat\tau_t^{\mathrm{DIM}}
\mid \mathcal F_{t-1}
\right]
=\tau$.
Consequently, adaptive covariate selection based only on previous
batches does not bias the batch-level design-based estimator.
\end{theorem}

Proof of Theorem \ref{thm:unbiased} is provided in Appendix \ref{appx:thmunbiased}; the argument is batch-local and does not require the covariate set to be fixed across batches.
Moreover, the combined across-batch estimator is asymptotically valid.

\begin{theorem}[Validity and conservativeness of the sequential ATE estimator]
\label{thm:conserv}
Define $\hat\mu_T$ and $\hat\sigma_T$ as above. 
Under sufficient high-level regularity conditions (stated in Appendix~\ref{app:regularity}) ensuring that estimated weights are equivalent to oracle inverse-variance weights (A1--A2); no single batch dominates the asymptotic variance (A3); rerandomization does not inflate variance relative to simple random assignment(A4--A5); and any conditional bias is asymptotically negligible (A6),

\[
\tfrac{\hat\mu_T-\tau}{\hat\sigma_T}
\xrightarrow{d}
\mathcal N(0,c^{-2})
\]
for some $c^2\ge1 $. Consequently,
$P\!\left(
|\hat\mu_T-\tau|
\le z_{\alpha/2}\hat\sigma_T
\right)
\to
P(|Z|\le z_{\alpha/2}c)
\ge 1-\alpha$, so the interval $\hat\mu_T\pm z_{\alpha/2}\hat\sigma_T $ has at least nominal asymptotic coverage.
\end{theorem}

Full proof can be found in Appendix \ref{appx:thmconserv}. 

While exactly unbiased under rerandomization, the difference-in-means estimator can be made more precise using regression adjustment \citep{li2020rerandomization}. 
For the Lin-adjusted estimator, the usual agnostic regression adjustment has a finite-sample bias of order $O(n^{-1})$ (see Appendix \ref{appx:asymptotic}). In the equal-batch-size case with, $\sigma_t^2\asymp n^{-1}$, and $\bar w_t\asymp n$, the accumulated bias is negligible relative to the standard error when $T=o(n)$. Thus ordinary Lin adjustment is covered in regimes where the batch size grows faster than the number of batches. More generally, the theorem requires only the stated negligible-bias condition.

The two inferential statements play different roles. The first is an exact design-based result: adaptive covariate selection does not bias the batch-level difference-in-means estimator because the selected set is fixed before the current assignment is randomized. The second is an asymptotic meta-analysis result: the cumulative estimator uses same-batch estimated precision weights, so exact unbiasedness is not available, but conservative coverage follows when those weights are consistent for a design-agnostic reference variance and the accumulated
centering bias is negligible.

\paragraph{Unknown and costly covariates.}
\label{sec:to_learn}
In practice, the most predictive covariates are rarely known in advance and costly to measure, forcing the experimenter to identify which variables are worth buying within a hard budget $B$ across all $T$ batches. The sequential structure offers a natural resolution: we \textit{learn} which covariates are most prognostic during the experiment, updating beliefs after each batch.

To measure importance, we consider efficiency gain from rerandomization and Lin's adjustment strategies, which scales with $R^2$, the proportion of variance in the potential outcomes explained by $S_t$ \citep{li2020rerandomization, morgan2012rerandomization}. 
More predictive covariates yield larger reductions in the standard error of the batch-level ATE estimator, making predictiveness the natural criterion for covariate selection. This motivates treating the prognostic importance of each covariate as the quantity to be learned: if covariates are independent, a more predictive covariate contributes more to $R^2$ and hence delivers greater efficiency gains. We evaluate the impact of this proxy in the empirical results.

\section{Variable selection via combinatorial semi-bandit with knapsacks}
We frame identification of prognostic factors problem as a sequential variable selection problem, following the approach of \citet{liu2023variable}, extending it to accommodate budget constraints and an explicit causal inference objective.

\subsection{Problem formulation}
We assume a known pool of $p $ pre-treatment candidate covariates, where each variable $j \in [p] = \{1, \dots, p\} $ has an associated measurement cost $c_j \in [0,1] $. These variables correspond to the set of arms available to the agent, and each arm $j$ has a stochastic reward with fixed but unknown distribution with mean $\theta^*_j \in [0,1] $. We are given a total budget $B$, and each round can select at most $K\leq p$ arms. 

At each time period $t \in [T] $, the agent selects a subset of arms $S_t \subset [p]$, observing a binary reward $r_j(t)\in \{0,1\} $ from each arm $j\in S_t $ with $\mathbb E[r_j(t)]=\theta_j^* $ (semi-bandit feedback).
The global reward of a super-arm $S_t $ is defined as $R(S_t) = \sum_{j \in S_t} r_j(t) $, with expected value $r_{\theta^*}(S_t) =\sum_{j \in S_t} \theta^*_j $.
Following \citet{liu2023variable}, we treat $\theta_j^* $ as a proxy for the prognostic importance of covariate $j $ for the control potential outcome $Y_i(0) $: an arm with high mean reward is one whose covariate is strongly predictive of baseline outcomes. The goal is to maximize the total expected reward subject to the overall budget constraint $B$. 

\subsection{From unconstrained to budget-constrained selection}
The oracle solution under no budget is to measure every covariate for every unit in every batch. In practice, of course, budget constraints make this infeasible, and the selection problem becomes non-trivial.
A natural attempt is to allocate a fixed per-batch budget $B/T $, spending it greedily within each round. \citet{das2022budgeted} show in Lemma 1, however, that per-batch budget schemes can perform arbitrarily badly: if the oracle set of covariates --- those with the highest prognostic value --- has total cost exceeding the per-batch budget, the algorithm will never be able to select them, and their rewards will never be observed regardless of how much total budget remains. 
A purely greedy strategy --- always pulling the highest-ranked arms first --- fails because early reward estimates are noisy: budget may be exhausted on arms that appeared promising before reliable rankings emerge, with no resources remaining to correct mistakes. While Explore-Then-Commit (ETC) is a natural alternative, its uniform exploration rapidly exhausts strict budgets on high-dimensional noise covariates. Because the optimal commit point relies on unknown parameters (gaps, noise, costs), ETC cannot dynamically truncate its search. Instead, it remains trapped in unguided exploration until budget depletion, yielding performance similar to a random allocation.

The resolution, proposed by \citet{badanidiyuru2018bandits} and extended by \citet{das2022budgeted}, is to pose the selection problem as the integer program (\ref{eq:problem}), relax it to a linear program, and work with the dual of that relaxation. Full problem formulation and extension following \citet{das2022budgeted} is presented in Appendix \ref{appx:lp}.

\subsection{Thompson sampling and regret bound}

Under binary rewards, arm $j $ yields $r_j(t) \in \{0,1\} $ with $\mathbb{E}[r_j(t)] = \theta^*_j $, placing us squarely in the Beta-Bernoulli conjugate setting. We maintain independent Beta posteriors $\theta_j \sim \text{Beta}(\alpha_j, \beta_j) $, initialized at $\alpha_j(0) = \beta_j(0) = 1 $ (unless prior knowledge is available) and update after each round via the standard conjugate update:
\begin{equation}
\label{eq:beta_update}
\alpha_j(t) \leftarrow \alpha_j(t-1) + r_j(t), \qquad \beta_j(t) \leftarrow \beta_j(t-1) + (1 - r_j(t)).
\end{equation}

At each round $t$, the idealized CBwK-LP-TS policy draws posterior samples $\tilde\theta_j(t) \sim \mathrm{Beta}(\alpha_j(t-1), \beta_j(t-1))$ for all $j \in [p]$ and uses these in place of the UCB estimates of \citet{das2022budgeted}, with all other aspects of the CBwK-LP selection procedure unchanged.
We call this algorithm CBwK-LP-TS, adapting the UCB-based CBwK-LP-UCB of \citet{das2022budgeted} to Thompson sampling.
The idealized policy assumes the LP dual solutions are known at each round, so that the only source of suboptimality is uncertainty about the true arm means $\theta^\star$. The Bayes learning regret of Theorem~\ref{thm:regret} below therefore measures exclusively the cost of substituting the posterior draw $\tilde\theta_t$ for $\theta^\star$ in the selection map.

\begin{theorem}[Bayes Risk of idealized CBwK-LP-TS] 
\label{thm:regret}
Let $\pi^{TS} $ denote the idealized CBwK-LP-TS algorithm with independent $\mathrm{Beta}(\alpha_j(0), \beta_j(0))$ priors on $\theta^*_j$.  Assume semi-bandit feedback with binary rewards $r_j(t) \in \{0,1\}$ satisfying  $\mathbb{E}[r_j(t) \mid \theta_j^*, \mathcal{F}_{t-1}] = \theta_j^* $, arms with costs $c_j \geq c_{\min}>0$, budget $B$, and horizon $T$. Define the regret $\mathrm{Regret}(T,\pi) = \sum_{t=1}^T [r_{\theta^*}(S^*_t) - r_{\theta^*}(S_t)]$. Then
\[
\mathrm{BayesRisk}(T, \pi^{TS}) \leq O\!\left(\sqrt{p\log(pT)\cdot \min\!\left(KT,\, \tfrac{B}{c_{\min}}\right)}\right),
\]
where $K \leq \min(p, \lfloor B/c_{\min} \rfloor) $ is the maximum feasible super-arm size.
\end{theorem}
Detailed proof can be found in the Appendix~\ref{appx:thmregret}.
Our proof leverages the Russo-Van Roy decomposition \citep{russo2014learning}, which bounds the Bayes risk using an artificial sequence of upper confidence bounds (UCBs). 
If the action map is an LP/primal-dual map with a known deterministic planning gap $A_{\mathrm{LP}}$ as defined by \citet{das2022budgeted}, the LP benchmark bound adds $\mathbb{E} [A_{\mathrm{LP}}]$ to the displayed learning term. Detailed discussion on planning gap is included in Appendix \ref{appx:planning}.

\paragraph{Lower bound analysis}
We establish a lower bound by considering a uniform-cost instance where arm identification must come entirely from reward observations, and applying the gap-free minimax lower bound of \citet{kveton2015tight} for combinatorial semi-bandits with linear rewards.
\begin{theorem}[Minimax Lower Bound]
\label{thm:lower_bound}
For any policy $\pi$, consider the regime with $K<p$,
$B \ge p\cdot c_{\min}$, $K\le \lfloor B/c_{\min}\rfloor$, and
$T \ge \lceil p/K\rceil$. Then there exists a Combinatorial Semi-Bandit
with Knapsacks instance with $p$ arms, costs $c_j\in(0,1]$, budget $B$,
and horizon $T$ such that
\[
\mathbb{E}[\mathrm{Regret}(T,\pi)] \geq \Omega\left(\sqrt{p \cdot \min\left(KT, \tfrac{B}{c_{\min}}\right)}\right).
\]

\end{theorem}
The conditions of the theorem ensure that the effective pull budget is at least $p$, placing the problem in the nontrivial regime where all arms could in principle be sampled and regret is driven by statistical identification rather than by an insufficient horizon or budget. Full proof is provided in Appendix \ref{appx:thm_lower}.
Theorem \ref{thm:regret} bounds the Bayes risk of CBwK-LP-TS via the framework of \citet{russo2014learning}, while Theorem~\ref{thm:lower_bound} establishes a frequentist minimax lower bound. These operate under different statistical frameworks: the Bayes risk averages over a prior on $\theta^* $, whereas the minimax bound holds for a specific hard instance. Nevertheless, in the non-trivial regime of the theorem, the polynomial terms match up to a $\sqrt{\log(pT)} $ factor. Establishing a formal frequentist upper bound for CBwK-LP-TS remains an avenue for future work.


\subsection{Practical considerations}
\paragraph{Predictiveness proxy and fractional rewards}
Efficiency gains from rerandomization and regression adjustment are typically measured using $R^2$. To avoid treatment effect contamination, we estimate predictiveness using control units only via LASSO regression of $Y(0,t)$.

Under standard sparsity conditions, LASSO provides consistent prediction in high dimensions \citep{meinshausen2009lasso, wasserman2009high}, and its coefficients serve as a proxy for prognostic importance \citep{lanners2023variable}. 

We define fractional rewards
\[
r_j(t) = \frac{|\hat{\lambda}_j|}{\max_{k \in S_t}|\hat{\lambda}_k|},
\]
with $r_j(t)=0$ if $\hat{\lambda}_j=0$. Treating $r_j(t)$ as pseudo-Bernoulli observations in a Beta update yields a pseudo-posterior that aggregates predictive signals over time, down-weighting variables 
that receive consistently low or unstable rewards.

\begin{wrapfigure}{r}{0.5\textwidth}
\vspace{-18pt}
  \begin{center}
  \includegraphics[width=0.48\textwidth]{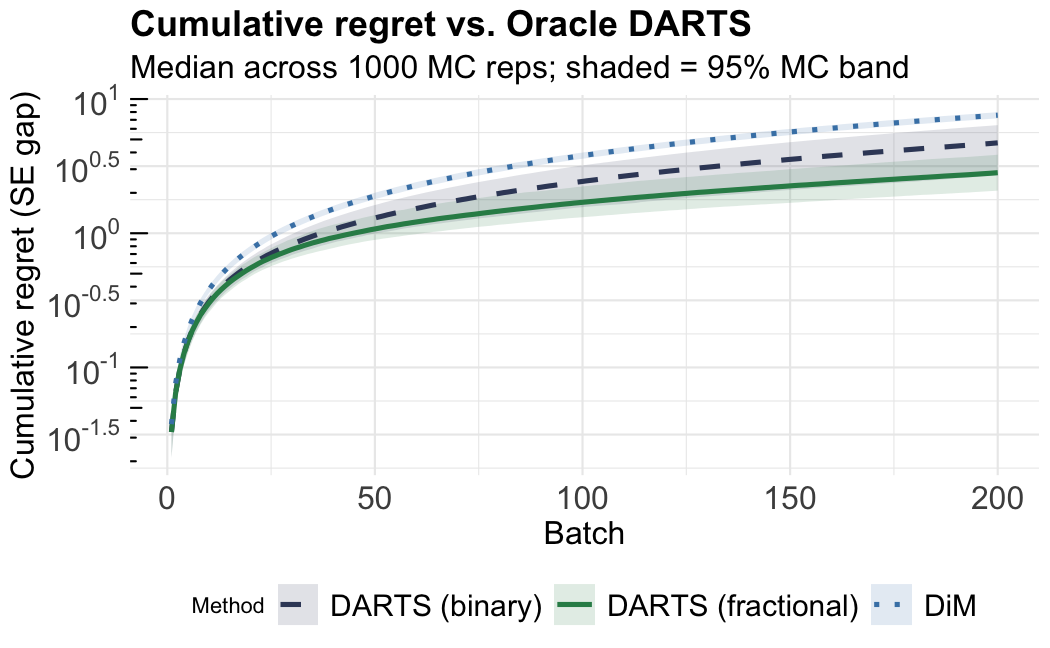}
  \caption{Cumulative regret relative to Oracle DARTS across 1000 
  replications. Solid lines show medians; shaded bands show 95\% 
  intervals from run percentiles. DARTS with fractional and binary 
  rewards exhibit sublinear growth, with fractional rewards 
  converging faster.}
  \label{fig:regret_bin_frac}
  \end{center}
  \vspace{-10pt}
\end{wrapfigure}
We formally characterize the regret bound of this fractional scheme in Appendix~\ref{appx:frac_rew}, establishing that the algorithm retains sublinear Bayes risk when the pseudo-posterior approximation error is sufficiently small. Such theory--implementation separation is common in posterior-sampling methods: regret analyses are often derived for tractable posterior updates, while practical systems use approximate or modified posterior-sampling rules whose behavior must be checked separately \citep{chapelle2011empirical, riquelme2018deep, qin2022analysis, phan2019thompson}. Empirically, as shown in Figure~\ref{fig:regret_bin_frac}, the fractional scheme achieves strictly lower regret than a binary reward update. It reduces regret relative to the oracle policy faster, motivating its use in the full DARTS algorithm.

\textbf{Note on independence.} The bandit framework assumes independent arm rewards. LASSO partially mitigates correlation violations by typically selecting a sparse set of representatives from collinear groups. Under a budget constraint, this is highly desirable: it prevents wasting budget on correlated substitutes while using fewer degrees of freedom without sacrificing predictive performance.

\paragraph{Back to Experimental Design.}
\label{sec:causal}

At each round $t $, once the super-arm $S_t $ is selected, the algorithm transitions from the bandit problem back to the causal inference problem. The selected covariates $\{X_{ij}(t) : j \in S_t\} $ are measured for each unit $i $ in batch $t $, yielding the covariate matrix $\mathbf{X}(t) \in \mathbb{R}^{n \times |S_t|} $, which is then used for treatment assignment and ATE estimation as described in Section \ref{sec:srt}.

\textbf{Outcome observation and ATE estimation.}
Once treatment is assigned, outcomes are observed and the batch-level ATE is estimated via Lin-adjusted OLS on the same 
 $|S_t| $ variables used for rerandomization, with HC2 heteroskedasticity-robust variance estimate $\hat{v}_t $, as described in Section~\ref{sec:srt}. No additional measurement cost is incurred. The cumulative estimate $\hat{\mu}_t $ is then updated via the precision-weighted meta-analysis in \eqref{eq:meta}.

Lin's regression adjustment fits separate within-arm regressions, consuming $2|S_t|$ additional parameters relative to the unadjusted estimator. When $|S_t|$ is large relative to $n$, OLS loses degrees of freedom and the HC2 variance estimate can become unstable. We therefore implicitly assume $|S_t| \ll n$ throughout. In practice this can be enforced by choosing larger batch sample sizes or additional budget constraint, which limits the number of covariates measured per batch. When the budget permits selecting many covariates, the analyst may wish to add an explicit degree-of-freedom constraint of the form $|S_t| \leq n/k$ for some $k > 2$ as an additional knapsack constraint in Problem \ref{eq:problem}. In our implementation we use the Moore-Penrose pseudoinverse to handle near-rank-deficient designs gracefully when estimating $\hat\Sigma(t)^{-1}$, but this is a numerical safeguard rather than a substitute for adequate sample size.

\textbf{Reward update.}
Finally, the observed batch is used to update the Thompson sampler. We fit a LASSO regression on the control units of batch $t $ and compute fractional rewards $r_j(t) = |\hat{\lambda}_j| / \max_k |\hat{\lambda}_k| $ for each $j \in S_t $, then apply the pseudo-posterior Beta update \eqref{eq:beta_update}. The updated posteriors $\{\mathrm{Beta}(\alpha_j(t), \beta_j(t))\}_{j=1}^p $ inform arm selection in round $t+1 $, closing the adaptive loop between variable selection and experimental design.

The complete procedure is summarized in Algorithm \ref{alg:DARTS}.

\begin{algorithm}
\caption{Dynamic Adaptive Rerandomization via Thompson Sampling (DARTS)}
\label{alg:DARTS}
\begin{algorithmic}[1]

\Require Covariate matrix $\mathbf{X} $, budget 
   $B $, costs $\{c_j\}_{j=1}^p $, batch size $n $, number of 
  batches $T $
\Ensure Sequential ATE estimates $\{\hat{\tau}_t\}_{t=1}^T $, 
  posterior inclusion probabilities $\{\pi_j\}_{j=1}^p $

\State \textbf{Initialise:} $\alpha_j \leftarrow 1 $, $\beta_j 
  \leftarrow 1 $, $\nu_j \leftarrow 1 $ for all $j \in [p] $; 
   $\nu_{\text{budget}} \leftarrow 1 $
\State Set $B_0 \leftarrow \min(B, T) $, $\tilde{c}_j \leftarrow 
  c_j \cdot B_0 / B $, $\varepsilon \leftarrow 
  \sqrt{\log(p+1)/B_0} $

\For{ $t = 1, \ldots, T $}

  \State \textbf{// Arm Selection (CBwK-LP-TS)}
  \State Sample $\tilde{\theta}_j \sim \mathrm{Beta}(\alpha_j, 
    \beta_j) $ for all $j \in [p] $
  \State Compute effective costs $\tilde{c}_j^{\text{eff}} 
    \leftarrow \tilde{c}_j \cdot \nu_{\text{budget}} + \nu_j $ 
    for all $j \in [p] $
  \State Select $S_t \subseteq [p] $ greedily by $\tilde{\theta}_j 
    / \tilde{c}_j^{\text{eff}} $, subject to remaining budget 
     $B_{t-1} $
  \State Update shadow prices: $\nu_j \leftarrow \nu_j(1+
    \varepsilon) $ and $\nu_{\text{budget}} \leftarrow 
    \nu_{\text{budget}}(1+\varepsilon)^{\tilde{c}_j} $ for all 
     $j \in S_t $
  \State $B_t \leftarrow B_{t-1} - \sum_{j \in S_t} \tilde{c}_j $

  \State \textbf{// Randomisation}
  \State Observe $\mathbf{X}(t) \in \mathbb{R}^{n \times |S_t|} $
  \State Draw $\mathcal{K} $ candidate assignments
  \State Set $Z(t) $ by 
    fair coin flip between $z^* $ and $\mathbf{1} - z^* $, where 
     $z^* = \arg\min_z\, d_M(z, \mathbf{X}(t)) $
  \State Observe $Y_i(t) = Z_i(t) Y_i(1) + (1-Z_i(t)) Y_i(0) $ 
    for all $i $

  \State \textbf{// Estimation}
  \State Fit $\hat{\tau}_t \leftarrow \hat{\beta}_Z $ from 
     regressing $(Y(t)$ on $Z(t),\mathbf{X}(t)$ and their interaction; 
     
    compute HC2 variance $\hat{v}_t $
  \If{ $t = 1 $}
    \State $\hat{\mu}_1 \leftarrow \hat{\tau}_1 $, \quad 
       $\hat{\sigma}_1^2 \leftarrow \hat{v}_1 $
  \Else
    \State $\hat{\sigma}_t^2 \leftarrow \bigl(\hat{\sigma}_{t-1}
      ^{-2} + \hat{v}_t^{-1}\bigr)^{-1} $, \quad 
       $\hat{\mu}_t \leftarrow \hat{\sigma}_t^2\bigl(
      \hat{\mu}_{t-1}/\hat{\sigma}_{t-1}^2 + 
      \hat{\tau}_t/\hat{v}_t\bigr) $
  \EndIf

  \State \textbf{// Thompson Sampler Update}
  \State Observe reward $r_j$ for all $j \in S_t $
  \State $\alpha_j \leftarrow \alpha_j + r_j $, \quad $\beta_j 
    \leftarrow \beta_j + (1-r_j) $ for all $j \in S_t $

\EndFor

\State \textbf{Return:} $\hat{\mu}_T $, $\hat{\sigma}_T $, 
   $\pi_j \leftarrow \alpha_j/(\alpha_j + \beta_j) $ for all 
   $j \in [p] $

\end{algorithmic}
\end{algorithm}

\section{Simulation Studies}
\label{sec:sims}
We evaluate DARTS across synthetic settings testing the learning mechanism, budget constraint, and inferential validity. Full reproducibility details are in Appendix~\ref{appx:sims}.

\subsection{Setup}
\label{sec:sims_setup}
\textbf{Covariates and outcome models.} We consider two DGPs with $q = 20$ truly predictive covariates out of $p \in \{100, 1000\}$ candidates. The \textbf{linear} DGP draws independent Gaussian covariates with a sparse linear outcome surface. The \textbf{Liang} DGP \citep{liang2018bayesian} introduces correlated covariates via a shared random effect and a nonlinear outcome, partially violating the arm-independence assumption. In both DGPs, $\tau = 4$.

\textbf{Budget and batch parameters.} The \textbf{Monte Carlo grid} uses $T = 100$ batches of $n \in \{100, 1000\}$ units, equal costs, and budgets $B \in \{1000, 2000, 20000\}$, averaged over 1000 replications. We compare \textbf{DiM} (complete randomization, no adjustment); \textbf{Random} (rerandomization and Lin adjustment on a randomly drawn budget-feasible covariate subset, no learning); \textbf{DARTS} (Algorithm~\ref{alg:DARTS}); and \textbf{Oracle} (rerandomization and Lin adjustment on the true $q = 20$ covariates, budget-free). The \textbf{comparison to ARMM and MADCovar} uses 1000 replications of the Liang DGP with $T = 200$ batches, $n = 1000$, $B = 2000$, and independent variable costs $c_j {\sim} \mathrm{Uniform}(0,2)$ --- a more difficult set up due to correlation and linearity violations. We report medians and 95\% bands from run percentiles. Two robustness checks with heterogeneous treatment effects and oracle-costly cost structures are in Appendix~\ref{appx:robustness}.

\subsection{Results}
\label{sec:sims_results}

\textbf{Rerandomization with adjustment.}
Table~\ref{tab:rerand-vs-adj} in the appendix confirms, across 1000 replications, that combining rerandomization with Lin adjustment consistently dominates rerandomization alone.
The advantage narrows when $p$ is large relative to $n$, as OLS adjustment consumes degrees of freedom, but never substantially harms precision. All subsequent comparisons use the combined estimator.

\begin{figure}[ht]
  \centering
  \begin{subfigure}[t]{0.48\textwidth}
    \centering
    \includegraphics[width=\textwidth]{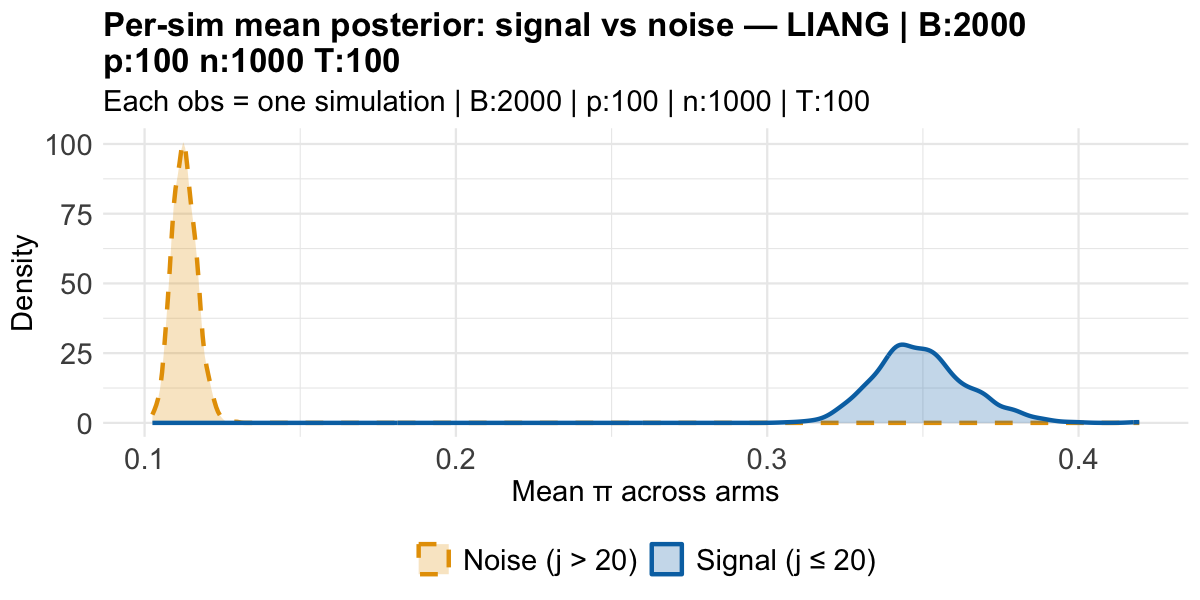}
    \caption{Liang, $B=2000$, $p=100$, $n=1000$: signal and noise
    distributions completely separated.}
  \end{subfigure}
  \hfill
  \begin{subfigure}[t]{0.48\textwidth}
    \centering
    \includegraphics[width=\textwidth]{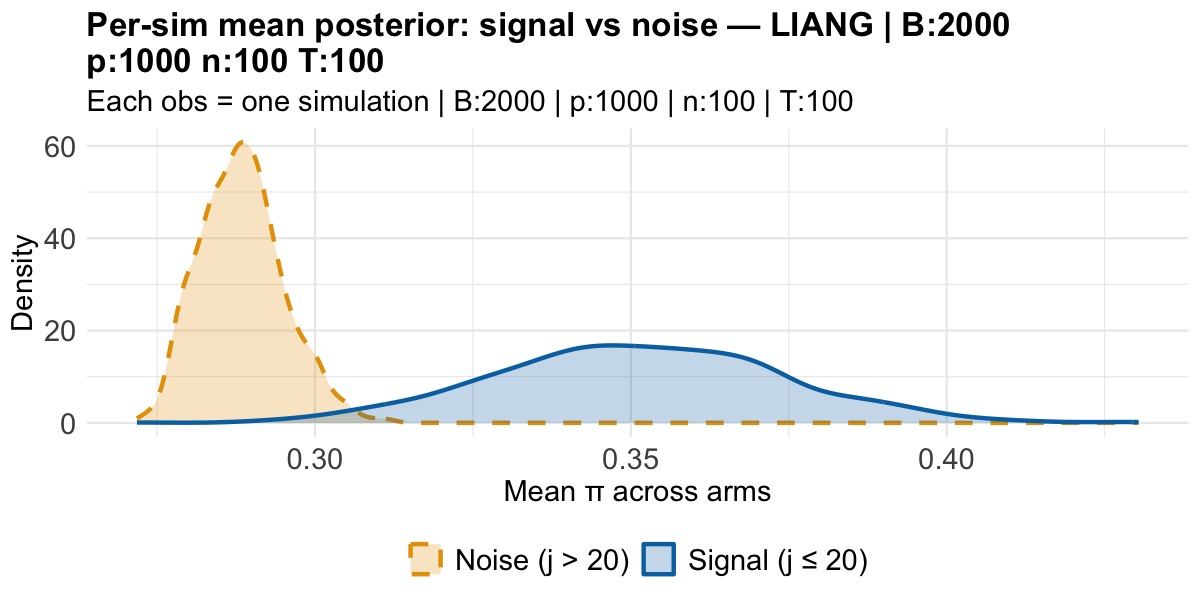}
    \caption{Liang, $B=2000$, $p=1000$, $n=100$: separation collapses
    as the learning problem becomes intractable.}
  \end{subfigure}
  \caption{Distribution of final posterior inclusion probabilities  $\pi_j = \alpha_j / (\alpha_j + \beta_j)$ across 1000 replications  for true signal ($j \leq 20$, solid blue) and noise ($j > 20$, dashed yellow) covariates. The Thompson sampler concentrates budget on prognostic covariates when the problem is tractable, and spreads it randomly 
  when the candidate pool is too large relative to the horizon.}
  \label{fig:posteriors}
\end{figure}

\textbf{Does DARTS learn the right covariates?} Figure~\ref{fig:posteriors} shows the distribution of final posterior inclusion probabilities $\pi_j = \alpha_j/(\alpha_j + \beta_j)$ across 1000 replications, separately for true signal ($j \leq 20$) and noise ($j > 20$) covariates. Under the Liang DGP with $B = 2000$, $p = 100$, $n = 1000$, the signal and noise distributions are completely separated (signal mean $0.34$, noise mean $0.11$), confirming that the Thompson sampler systematically concentrates budget on prognostic covariates. Separation is weaker under $B = 1000$ and $p = 1000$, where the sampler must identify 20 signal arms from 1000 candidates within a limited horizon, an honest boundary case where DARTS degrades gracefully toward the random baseline.
When budget is $20,000$ while $p=100$, essentially removing any monetary constraint, all methods perform similarly, since all get access to every oracle covariate.
Across the 1000-replication method comparison, the share of batch budget allocated to oracle variables increases monotonically over batches (Figure~\ref{fig:oracle_budget}, Appendix~\ref{appx:diagnostics}), confirming that the Thompson sampler progressively concentrates resources on the truly prognostic covariates as posterior beliefs sharpen with accumulated evidence.

\textbf{Does learning translate to precision gains?}
Table~\ref{tab:DARTS-vs-random} shows that $\mathrm{SE}(\mathrm{DiM}) \geq \mathrm{SE}(\mathrm{Random}) \geq \mathrm{SE}(\mathrm{DARTS}) > \mathrm{SE}(\mathrm{Oracle})$ holds for most DGPs and budgets. The main exceptions occur in high-$p$, low-$n$ settings, where the algorithm has insufficient data to learn effectively, and at high budgets, where all methods gain access to all oracle variables. DARTS achieves up to RE $= 3.56$ over DiM versus RE $= 1.26$ for Random, isolating the Thompson sampler's contribution. After a 30-batch burn-in, batch-level rewards correlate positively ($r = 0.977)$ with SE reduction relative to DiM (Figure~\ref{fig:reward_se},Appendix~\ref{appx:diagnostics}). Moreover, later batches correspond to higher SE reduction. This confirms that the LASSO fractional rewards proxy translates learning into precision gains. Coverage is at the nominal level throughout, consistent with Theorem~\ref{thm:conserv}.

\begin{figure}[ht]
  \centering
  \begin{subfigure}[t]{0.46\textwidth}
    \centering
    \includegraphics[width=\textwidth]{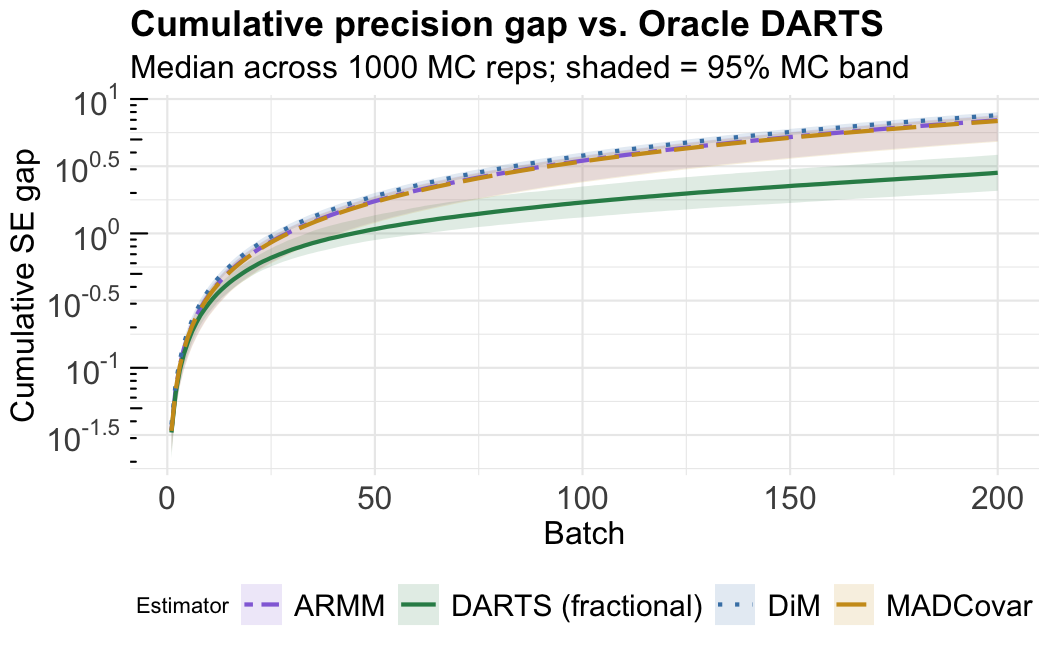}
    \caption{Cumulative regret relative to Oracle DARTS. Solid lines 
    show medians; shaded bands show 95\% 
    intervals from run percentiles.}
  \end{subfigure}
  \hfill
  \begin{subfigure}[t]{0.46\textwidth}
    \centering
    \includegraphics[width=\textwidth]{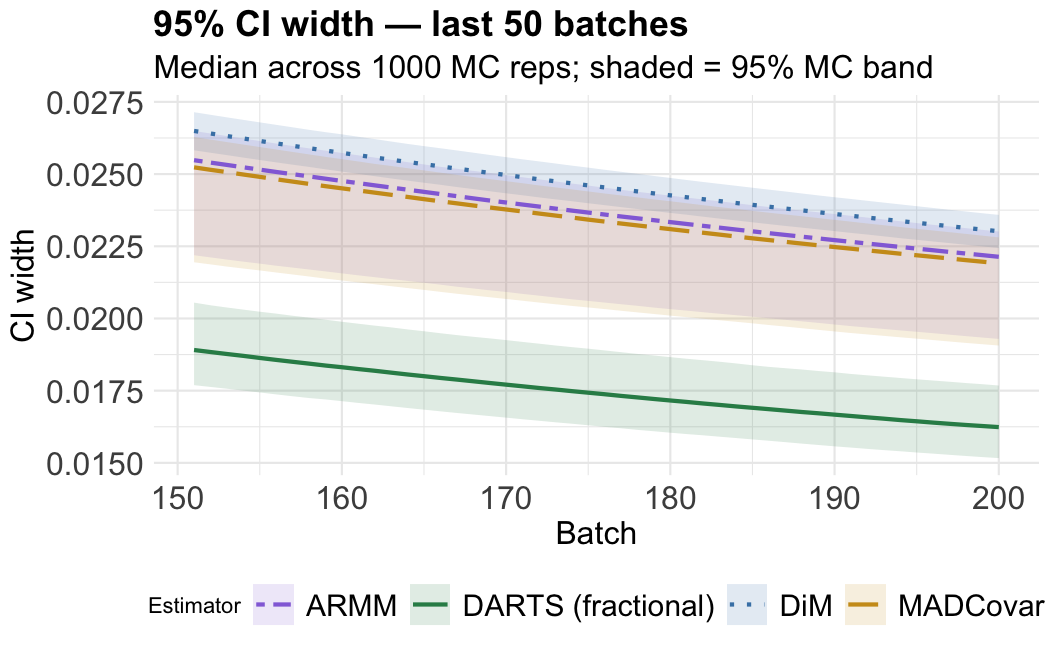}
    \caption{95\% CI width over the last 50 batches. Solid lines show 
    medians across 1000 replications; shaded bands show 95\% intervals 
    from run percentiles.}
  \end{subfigure}
  \caption{Comparison of DARTS against ARMM \citep{yang2023balancing} and MADCovar \citep{molitor2025anytime} across 1000 replications of the Liang DGP ($n = 1000$, $T = 200$, $p = 100$, $B = 2000$, variable costs). ARMM and MADCovar are given 12 randomly pre-selected covariates; DARTS learns from the full candidate pool. Solid lines show medians across 1000 replications; shaded bands show 95\% intervals from run percentiles. All CIs are 95\% valid at fixed $T$; HC2 standard 
  errors throughout.}
  \label{fig:baselines}
\end{figure}

\textbf{Comparison to ARMM and MADCovar.} \label{sec:madcovar_armm}
We compare against ARMM \citep{yang2023balancing} and MADCovar \citep{molitor2025anytime} over 1000 replications as described in Section~\ref{sec:sims_setup}. Both methods are given a randomly pre-selected subset of covariates per simulation --- drawn until the budget per batch is met, averaging around 12 --- the realistic scenario where an analyst pre-commits without knowledge of which are prognostic, while DARTS sees the full pool and must learn under the same budget.
Table~\ref{tab:dgp1-performance-mc} summarizes performance, where Rel.\ RMSE $= \mathrm{RMSE}_{\mathrm{DiM}} / \mathrm{RMSE}_{\mathrm{method}}$ with values above 1 indicating improvement over DiM. DARTS achieves Rel.\ RMSE $= 1.415$, compared to $1.018$ and $1.036$ for ARMM and MADCovar respectively, closing a substantial share of the gap to Oracle DARTS (Rel.\ RMSE $= 1.812$). 
Figure~\ref{fig:baselines} reinforces this: panel (a) shows that DARTS accumulates strictly less regret than ARMM and MADCovar throughout the horizon, and panel (b) shows that DARTS achieves narrower CI widths over the last 50 batches, with both differences remaining consistent across replications as seen by the non-overlapping 95\% bands.  The advantage of DARTS reflects the value of adaptive acquisition when the prognostic set is unknown --- when given oracle covariates, ARMM and MADCovar perform comparably to Oracle DARTS.

\section{Limitations and discussion}

\subsection{Limitations}
While DARTS consistently improves precision in unknown-covariate settings, we note several limitations regarding its reward proxy, theoretical guarantees, and implementation constraints. 

First, the theoretical regret guarantees rely on an idealized setting. Theorem~\ref{thm:regret} assumes exact Beta-Bernoulli conjugacy. While Corollary~\ref{cor:fractional} bounds the pseudo-posterior approximation error of fractional rewards, this bound depends on an unobservable total variation distance. The guarantee is robust in sparse settings but should be treated as indicative in dense-signal regimes. 

Second, the framework currently supports fixed-horizon inference. Theorem~\ref{thm:conserv} establishes asymptotic coverage only for the fixed-horizon estimator $\hat\mu_T$. The confidence intervals are valid only at the pre-specified end of the trial; sequential peeking or early stopping invalidates coverage. 

Finally, the empirical comparisons to ARMM and MADCovar are informationally asymmetric by design — DARTS learns from the full candidate pool, while baselines use randomly pre-selected covariate sets to accurately reflect standard practice. This asymmetry is inherent to the problem setting: ARMM and MADCovar are not designed to perform covariate acquisition and take a fixed covariate set as given, so the comparison is best understood as a demonstration of the value of adaptive acquisition over naive pre-selection rather than a head-to-head competition between methods solving the same problem.

\subsection{Discussion and future work}
DARTS demonstrates that it is possible to close the efficiency gap to oracle rerandomization without sacrificing inferential validity, gracefully degrading to a completely randomized baseline when the learning problem is intractable. 

To build on this framework, future work should directly address the limitations of the independent arm assumption. In settings where covariate clusters are known a priori, integrating alternative reward functions like group LASSO, or extending the framework to combinatorial bandits with correlation structure \citep{gupta2022correlated}, would allow the Thompson sampler to share information across correlated arms. Furthermore, extending the inverse-variance weighted estimator to utilize confidence sequence constructions would enable anytime-valid inference, a natural evolution for sequential and adaptive trial designs.

\bibliographystyle{plainnat}
\bibliography{references}


\appendix

\section{Technical appendices and supplementary material}
\subsection{Proof of theorem \ref{thm:unbiased}}
\label{appx:thmunbiased}

Fix batch $t $ and condition on $\mathcal F_{t-1} $. Since $S_t $ is chosen using only previous batches, the selected covariate set is fixed under this conditioning. Let
\[
\mathcal G_t
= \mathcal F_{t-1}\vee\sigma(\mathbf X(t))
\]
denote the sigma-field that additionally contains the selected covariates for batch $t $. Write $Y_i(1,t)=Y_i(0,t)+\tau_i(t) $, where $\tau_i(t)=Y_i(1,t)-Y_i(0,t) $. The unadjusted difference-in-means
estimator satisfies
\begin{align*}
\hat\tau_t^{\mathrm{DIM}}-\tau
&=
\underbrace{
\tfrac{1}{N_1(t)}\sum_{i:Z_i(t)=1}Y_i(0,t)
-
\tfrac{1}{N_0(t)}\sum_{i:Z_i(t)=0}Y_i(0,t)
}_{\mathrm{(I)}} \\
&\quad+
\underbrace{
\tfrac{1}{N_1(t)}\sum_{i:Z_i(t)=1}\tau_i(t)-\tau
}_{\mathrm{(II)}} .
\end{align*}

Conditional on $\mathcal G_t $, the selected covariates and potential outcomes in batch $t $ are fixed. By the mirror-symmetric rerandomization rule,
\[
P_t(Z(t)=z\mid\mathcal G_t)
=
P_t(Z(t)=\mathbf 1-z\mid\mathcal G_t).
\]
Thus the treated and control labels are exchangeable under the conditional randomization distribution. Therefore
\[
\mathbb E[\mathrm{(I)}\mid \mathcal G_t]=0.
\]

The same symmetry implies that the treated units are, in expectation under the conditional randomization distribution, representative of the finite batch with respect to the individual treatment effects:
\[
\mathbb E\!\left[
\tfrac{1}{N_1(t)}
\sum_{i:Z_i(t)=1}\tau_i(t)
\middle| \mathcal G_t
\right]
=
\bar\tau_t,
\]
where $\bar\tau_t=n_t^{-1}\sum_{i=1}^{n}\tau_i(t) $ is the finite-batch
average treatment effect. Hence
\[
\mathbb E[\mathrm{(II)}\mid \mathcal G_t]=\bar\tau_t-\tau.
\]
Taking expectations over the i.i.d. sampling of the batch, which is independent of $\mathcal F_{t-1} $, gives
\[
\mathbb E[\bar\tau_t\mid\mathcal F_{t-1}]=\tau.
\]
By iterated expectation,
\[
\mathbb E[
\hat\tau_t^{\mathrm{DIM}}-\tau
\mid \mathcal F_{t-1}
]
=0,
\]
which proves the claim.


\subsection{Asymptotic unbiasedness under covariate adjustment}
\label{appx:asymptotic}

Under the rerandomization procedure and regularity conditions (i)–(iii) below, the Lin-adjusted estimator satisfies $|\mathbb{E}[e_t \mid \mathcal{F}_{t-1}]| = O(n^{-1}) $, where $n $ is the batch size.

\textbf{Regularity conditions:} (i) Covariates are uniformly bounded: $\|X_i\| \leq C < \infty $. (ii) The finite population covariance matrices $n_z^{-1}\sum_{i:Z_i=z} X_i X_i^\top $ converge to positive definite limits $\Sigma_z $ for $z \in \{0,1\} $. (iii) Potential outcomes satisfy $\mathbb{E}[|Y_i(z)|^2] < \infty $.

To simplify notation, we assume without loss of generality that the covariates have already been centered within each batch such that $\bar{X}_t = \mathbf{0} $. The Lin-adjusted estimator is defined as:
\[
\hat{\tau}_{t,\mathrm{Lin}} = (\bar{Y}_{t,1} - \hat{\beta}_{t,1}^\top \bar{X}_{t,1}) - (\bar{Y}_{t,0} - \hat{\beta}_{t,0}^\top \bar{X}_{t,0}),
\]
where $\hat{\beta}_{t,z} $ are arm-specific OLS coefficients. We decompose the error $e_t = \hat{\tau}_{t,\mathrm{Lin}} - \tau $ as follows:
\begin{align*}
e_t &= \underbrace{(\bar{Y}_{t,1}(0) - \beta_1^{*\top}\bar{X}_{t,1}) - (\bar{Y}_{t,0}(0) - \beta_0^{*\top}\bar{X}_{t,0})}_{\text{(I)}} \\
&\quad + \underbrace{\bar{\tau}_{t,1} - \tau}_{\text{(II)}} \\
&\quad + \underbrace{(\beta_1^* - \hat{\beta}_{t,1})^\top \bar{X}_{t,1} - (\beta_0^* - \hat{\beta}_{t,0})^\top \bar{X}_{t,0}}_{\text{(III)}},
\end{align*}
where $\beta_z^* $ denote the population limits of the regression coefficients that exists under conditions (ii) and (iii).

The expectations of Terms (I) and (II) vanish exactly following the symmetry and sampling arguments established in the rerandomization proof (\ref{appx:thmunbiased}).
By the marginal $1/2 $ symmetry of the mirror-flip procedure, $P(Z_i(t)=1 \mid \mathbf X (t), \mathcal{F}_{t-1}) = 1/2 $. Given that potential outcomes and $\beta_z^* $ are fixed conditional on $\mathbf X (t) $, the expected adjusted baseline outcomes are equal across arms, so $\mathbb{E}[\text{(I)} \mid \mathbf X (t), \mathcal{F}_{t-1}] = 0 $. 
By the marginal $1/2 $ symmetry, $\mathbb{E}[\bar{\tau}_{t,1} \mid \mathbf X (t), \mathcal{F}_{t-1}] = \bar{\tau}_t $, the batch average. Taking the outer expectation over the i.i.d. sampling of units gives $\mathbb{E}[\bar{\tau}_t \mid \mathcal{F}_{t-1}] = \tau $, thus $\mathbb{E}[\text{(II)} \mid \mathcal{F}_{t-1}] = 0 $.

Term (III) captures the finite-sample bias of the OLS adjustment. Under condition (ii), the coefficient estimation error satisfies $\mathbb{E}[\|\hat{\beta}_{t,z} - \beta_z^*\|^2 \mid \mathcal{F}_{t-1}] = O(n^{-1}) $. Under rerandomization and within-batch centering, the group-mean imbalance satisfies $\mathbb{E}[\|\bar{X}_{t,z}\|^2 \mid \mathcal{F}_{t-1}] = O(n^{-1}) $. Applying the Cauchy-Schwarz inequality in $L^2 $:
\begin{align*}
\mathbb{E}\left[|(\beta_z^* - \hat{\beta}_{t,z})^\top \bar{X}_{t,z}| \middle|\mathcal{F}_{t-1}\right] & \leq \mathbb{E}[\|\hat{\beta}_{t,z} - \beta_z^*\|^2 \mid \mathcal{F}_{t-1}]^{1/2} \cdot \mathbb{E}[\|\bar{X}_{t,z}\|^2 \mid \mathcal{F}_{t-1}]^{1/2}\\
&= O(n^{-1}).
\end{align*}
Therefore, $|\mathbb{E}[\text{(III)} \mid \mathcal{F}_{t-1}]| = O(n^{-1}) $.
Combining these results, the sequence $\{e_t\} $ satisfies $|\mathbb{E}[e_t \mid \mathcal{F}_{t-1}]| = O(n^{-1}) $.

\subsection{Regularity conditions for Theorem~\ref{thm:conserv}}
\label{app:regularity}

As before, for each batch $t$, let
\[
\mathcal G_t
=
\mathcal F_{t-1}\vee\sigma(\mathbf{X}(t))
\]
be the sigma-field containing previous batches and the selected covariates for the current batch. Let $\hat\tau_t $ be the batch-specific ATE estimator and let $e_t=\hat\tau_t-\tau $. Define
\[
\hat\mu_T
= 
\tfrac{\sum_{t=1}^T w_t\hat\tau_t}{\sum_{t=1}^T w_t},
\qquad
\hat\sigma_T^2
=
\left(\sum_{t=1}^T w_t\right)^{-1},
\qquad
w_t=\hat v_t^{-1}.
\]
Let
\[
\sigma_t^2
=
\operatorname{Var}(\hat\tau_t\mid\mathcal G_t)
\]
denote the conditional variance under the actual rerandomization design, and let
\[
\sigma_{t,\circ}^2
=
\operatorname{Var}_{\circ}(\hat\tau_t\mid\mathcal G_t)
\]
denote the corresponding design-agnostic variance under simple random assignment. Set $\bar w_t=\sigma_{t,\circ}^{-2} $. Let
\[
V_T
=
\tfrac{\sum_{t=1}^T\bar w_t^2\sigma_t^2}
     {\left(\sum_{t=1}^T\bar w_t\right)^2}.
\]
We impose the following conditions.
\begin{enumerate}
    \item[A1] (Relative weight consistency) $ \max_{t\le T}
    \left|
    \tfrac{w_t}{\bar w_t}-1
    \right|
    \xrightarrow{p}0$.
    \item[A2] (Estimated-weight remainder) $\sum_{t=1}^T
    (w_t-\bar w_t)e_t
    =
    o_p\!\left(
    \sqrt{\sum_{t=1}^T\bar w_t^2\sigma_t^2}
    \right)$.
    \item[A3] (Uniform asymptotic negligibility) $\max_{t\le T}
    \tfrac{\bar w_t^2\sigma_t^2}
         {\sum_{s=1}^T\bar w_s^2\sigma_s^2}
    \to0.$
    \item[A4] (Variance ratio convergence) $\dfrac{\sum_t \bar{w}_t^2\sigma_t^2}{\sum_t\bar{w}_t} \xrightarrow{p} \ell \in (0,1]$
    
    \item[A5] (Conservative reference variance) $\sigma_t^2\le \sigma_{t,\circ}^2
    \quad\text{for all }t$.
    \item[A6] (Negligible centering bias) $ \sum_{t=1}^T
    \bar w_t\,
    \mathbb E[e_t\mid\mathcal G_t]
    =
    o_p\!\left(
    \sqrt{\sum_{t=1}^T\bar w_t^2\sigma_t^2}
    \right)$.
    \item[A7] (Bounded moments) There exist $\delta>0$ and $M<\infty$ such that $\sup_t \mathbb{E}[|e_t|^{2+\delta}\mid\mathcal{G}_t]\le M.$
\end{enumerate}
Condition~(A3) is a uniform asymptotic negligibility (UAN) condition for the martingale difference triangular array $\{\bar{w}_t(e_t-\mathbb{E}[e_t\mid\mathcal{G}_t])\}$. 
Combined with the bounded moments Condition~(A7), these jointly imply the conditional Lindeberg condition required by the Martingale CLT.
Conditions (A1)--(A2) permit replacement of $w_t$ with $\bar w_t$. 
Condition (A6) rules out asymptotically non-negligible centering bias. 
Conditions (A4)--(A5) ensure conservative variance estimation under rerandomization. (A5) is satisfied whenever the within-batch assignment mechanism is a rerandomization scheme in the sense of \citet{morgan2012rerandomization}. In the finite-population design-conditional setting, \citet{li2020rerandomization} show that such schemes reduce or preserve variance relative to simple random assignment. (A5) imports this conclusion into the superpopulation framework as a primitive assumption on the joint conditional distribution of $\hat{\tau}$ given $\mathcal{G}_t$.

\subsection{Proof of theorem \ref{thm:conserv}}
\label{appx:thmconserv}

By relative weight consistency,
\[
\tfrac{\sum_{t=1}^T w_t}{\sum_{t=1}^T\bar w_t}
\xrightarrow{p}1.
\]
By the estimated-weight remainder condition,
\[
\sum_{t=1}^T(w_t-\bar w_t)e_t
=
o_p\!\left(
\sqrt{\sum_{t=1}^T\bar w_t^2\sigma_t^2}
\right).
\]
Therefore,
\[
\hat\mu_T-\tau
=
\tfrac{\sum_{t=1}^T\bar w_t e_t}
     {\sum_{t=1}^T\bar w_t}
+o_p(\sqrt{V_T}).
\]

By the negligible centering-bias condition,
\[
\sum_{t=1}^T
\bar w_t\,
\mathbb E[e_t\mid\mathcal G_t]
=
o_p\!\left(
\sqrt{\sum_{t=1}^T\bar w_t^2\sigma_t^2}
\right).
\]
Thus the centered triangular array
\[
\bar w_t
\left\{
e_t-\mathbb E[e_t\mid\mathcal G_t]
\right\}
\]
is a martingale difference array with respect to $\{\mathcal{G}_t\}$ with sum of conditional variances equal to $\sum_{t=1}^T\bar w_t^2\sigma_t^2$ by construction.
To verify the Lindeberg condition, note that for any $\varepsilon>0$, by Markov's inequality and Condition~(A7):
\begin{align*}
&\sum_{t=1}^T \mathbb{E}\!\left[\frac{\bar w_t^2\{e_t-\mathbb{E}[e_t\mid\mathcal{G}_t]\}^2}{\sum_s \bar w_s^2\sigma_s^2}
\cdot
\mathbf{1}\!\left(\frac{|\bar w_t\{e_t-\mathbb{E}[e_t\mid\mathcal{G}_t]\}|}{\sqrt{\sum_s\bar w_s^2\sigma_s^2}}
>\varepsilon \right)
\,\Bigg|\,\mathcal{G}_t
\right]\\
&\le\frac{2^{1+\delta}M}{\varepsilon^\delta}
\cdot
\frac{\sum_{t=1}^T\bar w_t^{2+\delta}}{\left(\sum_s\bar w_s^2\sigma_s^2\right)^{1+\delta/2}}\to 0,
\end{align*}
where the last step follows from conditions~(A3) and~(A7).
The Lindeberg condition is therefore satisfied.
By~(A6) and Slutsky's theorem, the centering bias is negligible, and the Martingale CLT \citep{hall2014martingale} gives:
\[
\tfrac{\sum_{t=1}^T\bar w_t e_t}
     {\sqrt{\sum_{t=1}^T\bar w_t^2\sigma_t^2}}
\xrightarrow{d}
\mathcal N(0,1).
\]
Equivalently,
\[
\tfrac{\hat\mu_T-\tau}{\sqrt{V_T}}
\xrightarrow{d}
\mathcal N(0,1).
\]

Let
\[
\sigma_{T,\circ}^2
=
\left(\sum_{t=1}^T\bar w_t\right)^{-1}.
\]
Because $\bar w_t=\sigma_{t,\circ}^{-2}$ and using (A5),
\[
\tfrac{V_T}{\sigma_{T,\circ}^2}
=
\tfrac{\sum_{t=1}^T\bar w_t^2\sigma_t^2}
      {\sum_{t=1}^T\bar w_t}
\le
\tfrac{\sum_{t=1}^T\bar w_t^2\sigma_{t,\circ}^2}
      {\sum_{t=1}^T\bar w_t}
=1.
\]
By Condition~(A4), $V_T/\sigma_{T,\circ}^2\xrightarrow{p} c^{-2}$, where $c^2 = \ell^{-1}\ge1$. Relative
weight consistency also yields
\[
\tfrac{\hat\sigma_T^2}{\sigma_{T,\circ}^2}
=
\tfrac{\left(\sum_t w_t\right)^{-1}}
     {\left(\sum_t \bar w_t\right)^{-1}}
\xrightarrow{p}1.
\]
By Slutsky's theorem,
\[
\tfrac{\hat\mu_T-\tau}{\hat\sigma_T}
\xrightarrow{d}
\mathcal N(0,c^{-2}).
\]
Since $c^{-2}\le1$, the Wald interval based on $\hat\sigma_T$ has
asymptotic coverage at least $1-\alpha$.


\subsection{LP dual problem and shadow prices}
\label{appx:lp}
The integer programming program can be formulated as follows:
\begin{equation}
\label{eq:problem}
\begin{aligned}
\max_{a(1), \dots, a(T)} \quad & \sum_{t=1}^T \sum_{j=1}^p 
  a_j(t)\theta^*_j \\
\text{subject to} \quad 
  & \sum_{t=1}^T \sum_{j=1}^p a_j(t)c_j \leq B, 
  & \sum_{t=1}^T a_j(t) \leq T, \quad \quad a_j(t) \in \{0,1\} \quad \forall j \in [p], t \in [T],
\end{aligned}
\end{equation}
where $a_j(t)=1$ indicates covariate $j$ is measured in batch $t$.
The mean rewards $\theta^*_j $'s are unknown and must be learned sequentially, compounding the difficulty of Problem \eqref{eq:problem} beyond its NP-hard nature.

The LP relaxation replaces the binary constraint $a_j(t) \in \{0,1\} $ with a continuous allocation variable $\zeta_j \geq0 $:
\begin{equation}
\label{eq:lp}
\max_{\zeta \geq 0} \sum_{j=1}^p \zeta_j \theta^*_j \quad \text{subject to} \quad \sum_{j=1}^p \zeta_j M_{ij} \leq B' \quad \forall\, i \in \{1, \dots, p+1\},
\end{equation}
where $B' = \min(B,T) $, $\tilde{c}_j = c_j \cdot B'/B $ are rescaled costs, and $M $ encodes both the global budget and per-arm frequency constraints on a common scale. By Lemma 3 of \citet{das2022budgeted}, $\mathrm{OPT}_{LP} \geq \mathrm{OPT}_{IP} $, so benchmarking against $\mathrm{OPT}_{LP} $ provides a valid upper bound on the integer optimum.

\paragraph{Shadow prices and the dual problem}
Rather than solving \eqref{eq:lp} at each round, we work with its dual. The dual of \eqref{eq:lp} introduces a shadow price $\nu_i \geq 0 $ for each resource constraint $i$. Intuitively, a shadow price measures how much additional reward one extra unit of budget would yield: strictly positive when the constraint is binding, zero when slack.
We maintain a vector of dual shadow prices $\nu \in \mathbb{R}^{p+1} $, one per arm and one for the global budget constraint, initialized to 1 and updated multiplicatively after each round:
\[
\nu_j \leftarrow \nu_j(1 + \varepsilon), \qquad \nu_{p+1} \leftarrow \nu_{p+1}(1 + \varepsilon)^{\tilde{c}_j} \quad \forall j \in S_t,
\]
where $\varepsilon = \sqrt{\log(p+1)/B_0} $ is the multiplicative weight learning rate. Each time an arm is pulled, its shadow price increases, making it less attractive in subsequent rounds. The update has a natural interpretation: having spent heavily on arm $j $ in recent rounds, the algorithm treats the remaining budget as more valuable and demands a higher return per unit cost before pulling $j $ again. On average, this prevents early over-spending and ensures the global budget is not breached. The effective cost of arm $j $ at round $t $ is then:
\[
\tilde{c}_j^{\text{eff}}(t) = \tilde{c}_j \cdot \nu_{p+1}(t) + \nu_j(t),
\]
which inflates the nominal cost of frequently-pulled arms. Looking ahead over the remaining $H = T - t + 1 $ rounds, the super-arm is selected greedily by the bang-per-buck ratio:
\[
S_t = \mathrm{Greedy}\!\left(\left\{\tfrac{\tilde{\theta}_j(t)}{\tilde{c}_j^{\text{eff}}(t)}\right\}_{j \in [p]},\; B_{t-1}\right),
\]
where $\mathrm{Greedy} $ sorts arms in decreasing order of $\tilde{\theta}_j / \tilde{c}_j^{\text{eff}} $
and selects greedily subject to the remaining budget $B_{t-1} $.


\subsection{Proof of theorem \ref{thm:regret}}
\label{appx:thmregret}

The Bayes risk over $T$ periods is defined as the expected cumulative regret against the optimal subset $S^*_t$:
$$BayesRisk(T,\pi^{TS}) = \sum_{t=1}^{T}\mathbb{E}[r_{\theta^*}(S_t^*) - r_{\theta^*}(S_t)],$$
where the expectation is over the prior on $\theta^*$ and the algorithm's randomness. Let $\mathcal{F}_{t-1}$ be the filtration generated by the implemented Beta policy, $\pi^{TS}$, up to round $t$.

To analyze the Thompson Sampling policy without directly tracking the posterior quantiles, we apply the fundamental identity of Russo and Van Roy \citep{russo2014learning} (Proposition 1). Because Thompson Sampling ensures that the distribution of the selected arm $S_t$ matches the posterior distribution of the optimal arm $S^*_t$, we can introduce any valid $\mathcal{F}_{t-1}$-measurable upper confidence bound sequence $\{U_t \mid t \in \mathbb{N}\}$ as an analytical tool. We can then decompose the Bayes Risk into a confidence width term (Term I) and a UCB failure term (Term II):
\begin{align*}
BayesRisk(T,\pi^{TS}) &= \mathbb{E}\sum_{t=1}^{T}[U_t(S_t) - r_{\theta^*}(S_t)] + \mathbb{E}\sum_{t=1}^{T}[r_{\theta^*}(S_t^*) - U_t(S_t^*)].\\
&= \text{Term I} + \text{Term II}
\end{align*}

Since we have semi-bandit feedback, we can construct independent confidence bounds. Let $\hat\theta_{j}(t) = \tfrac{1}{n_j(t)}\sum_{s<t, j\in S_s}r_j(s) $ (with $\hat\theta_j(t) = 0 $ if $n_j(t) = 0 $), where $n_j(t) $ denotes the number of times arm $j $ was pulled up to time $t $. For each $j\in [p] $ and round $t $, we define:
\[
  U_{j,t} = \min \left\{1, \hat\theta_{j}(t) + \sqrt{\tfrac{\log(pT^2)}{2n_{j}(t)}}\right\}, \qquad L_{j,t} = \max\left\{0, \hat\theta_{j}(t) - \sqrt{\tfrac{\log(pT^2)}{2n_{j}(t)}}\right\},
\]
with $U_{j,t} = 1 $ and $L_{j,t} = 0 $ when $n_j(t) = 0 $.

Since $r_j(t) \in [0,1] $ are bounded, by Hoeffding's inequality (assuming $\mathbb{E}[r_j(t) \mid \theta_j^*, \mathcal{F}_{t-1}] = \theta_j^* $) for each arm $j $ at each round $t $:
\begin{align*}
  P(\theta_j^* > U_{j,t}\mid \mathcal{F}_{t-1}) &= P\left(\theta_j^* > \hat\theta_{j,t} + \sqrt{\tfrac{\log(pT^2)}{2n_j(t)}}\middle| \mathcal F_{t-1}\right) \\
  &= P\left(\theta_j^* -\hat\theta_{j,t} >\sqrt{\tfrac{\log(pT^2)}{2n_j(t)}}\middle| \mathcal F_{t-1}\right) \\
  &\leq \exp\left\{-\tfrac{2 n_{j}(t) \log(pT^2)}{2n_j(t)}\right\}\\
  &= \tfrac{1}{pT^2},
\end{align*}
and, symmetrically $P(\theta_k^* < L_{j,t} \mid\mathcal F_{t-1}) \leq 1/(pT^2) $. We define the super-arm bounds additively by linearity of the reward:
\[
U_t(S) =\sum_{j\in S} U_{j,t}, \quad 
L_t(S) = \sum_{j\in S} L_{j,t}.\] 
\textbf{Term II bound:} The event $r_{\theta^*}(S_t^*) > U_t(S_t^*) $ requires at least one arm $j\in S_t^* $ to satisfy $\theta_j^* > U_{j,t} $. By the union bound over at most $K $ arms in $S_t^* $:
\[
P(r_{\theta^*}(S_t^*) > U_t(S_t^*)\mid \mathcal{F}_{t-1}) \leq \sum_{j\in S_t^*}P(\theta_j^* > U_{j,t} \mid \mathcal{F}_{t-1}) \leq K\tfrac{1}{pT^2}.
\]
Since each arm contributes at most 1 to the reward, $r_{\theta^*}(S_t^*) - U_t(S_t^*) \leq |S_t^*| \leq K $. Thus,
\[
\mathbb{E}\sum_{t=1}^T [r_{\theta^*}(S_t^*) - U_t(S_t^*)] \leq \sum_{t=1}^T K \tfrac{K}{pT^2} = \tfrac{K^2}{pT}.
\]
This term vanishes as $T\to \infty $.

\textbf{Term I bound:} 
Let $\mathcal{E}_t = \{L_{j,t} \leq \theta_j^\star \leq U_{j,t} \ \forall j\in [p]\} $ denote the event that all confidence bounds hold at round $t $. 
By union bound and Hoeffding's inequality:
\begin{align*}
  P(\mathcal{E}_t^C \mid \mathcal{F}_{t-1}) &= P(\exists j : \theta_j^* < L_{j,t} \vee \theta_j^* > U_{j,t}\mid \mathcal{F}_{t-1}) \\
  &\leq \sum_{j=1}^p P(\theta_j^* < L_{j,t} \vee \theta_j^* > U_{j,t}\mid \mathcal{F}_{t-1})\\
  &\leq p \tfrac{2}{pT^2} = \tfrac{2}{T^2}.
\end{align*}
This complementary event contributes at most $K\cdot 2/T^2 $ per round in expectation. Thus, the total expected penalty is at most $ \sum_{t=1}^T \tfrac{2K}{T^2} = \tfrac{2K}{T^2} \cdot T = \tfrac{2K}{T} = O(K/T). $

Therefore, on high probability event $\mathcal{E}_t $:
\[
U_t(S_t) - r_{\theta^*}(S_t) \leq U_t(S_t) - L_t(S_t) = \sum_{j\in S_{t}}(U_{j,t} - L_{j,t}) = \sum_{j\in S_t} 2\sqrt{\tfrac{\log(pT^2)}{2n_j(t)}}.
\]
Because semi-bandit feedback increments $n_j(t) $ independently for each arm $j $, we can exchange the order of summations over time and arms.
\begin{align*}
  \mathbb E\sum_{t=1}^T[U_t(S_t) - r_{\theta^*}(S_t)] &\leq \sum_{t=1}^T\sum_{j\in S_t} 2\sqrt{\tfrac{\log(pT^2)}{2n_{j}(t)}} &\\
  &= \sqrt{2\log(pT^2)}\sum_{j=1}^p\sum_{t=1}^T \tfrac{1}{\sqrt{n_j(t)}}& \\
  &\leq \sqrt{2\log(pT^2)}\sum_{j=1}^p 2\sqrt{n_j(T)} &\text{harmonic sum bound}\\
  &= 2\sqrt{2\log(pT^2)} \sqrt{\left(\sum_{j=1}^p \sqrt{n_j(T)}\right)^2} \\
  & \leq 2\sqrt{2\log(pT^2)}\sqrt{p\sum_{j=1}^p n_j(T)} & \text{Cauchy-Schwarz inequality}\\
  &= 2\sqrt{2\log(pT^2)} \sqrt{p Q(T)},
\end{align*}
where $Q(T) = \sum_{j=1}^p n_j(T) $ is the total number (quantity) of arms pulled across all rounds and arms.
The algorithm enforces two physical constraints on the number of arms that can be pulled:
\begin{enumerate}
  \item at most $K $ arms per found for $T $ rounds, so $Q(T) \leq KT $;
  \item each pull of arm $j $ costs at least $c_{min} $, and the total budget is $B $, so $c_{min} \cdot Q(T) \leq B $, giving $Q(T) \leq B/c_{min} $.
\end{enumerate}
Since both must hold, $Q(T)\leq \min\left(KT, \tfrac{B}{c_{min}}\right) $.
We get 
\[
 \mathbb E\sum_{t=1}^T[U_t(S_t) - r_{\theta^*}(S_t)] \leq 2\sqrt{2\log(pT^2)}\cdot\sqrt{p\cdot\min\left(KT, \tfrac{B}{c_{min}}\right)}.
\]
Combining this with the vanishing terms for event $\mathcal{E}_t^C $ and term II, we get the bound in the theorem.

\subsection{Planning gap}
\label{appx:planning}
Let $B' = \min(B, T)$ and let $L(\theta^\star) = \mathrm{OPT}_{LP}(\theta^\star)$ denote the LP  optimal value under true means $\theta^\star$. Following \citet{das2022budgeted}, the deterministic planning gap of the CBwK-LP primal-dual policy under known means is
\[
A_{\mathrm{Das}}(T, \theta^\star) 
= C_D \, L(\theta^\star) 
  \left( 
    \sqrt{\frac{\log (p+1)}{B'}} + \frac{p}{B'} 
  \right),
\]
where $C_D > 0$ is a universal constant. This bounds how far the primal-dual policy $\pi^{LP}(\theta^\star)$ falls short of $\mathrm{OPT}_{LP}(\theta^\star)$ when arm means are known, capturing the resource-allocation loss from online dual updates rather than statistical uncertainty about $\theta^\star$. Taking expectations over the prior gives
\[
\mathrm{Gap}_{LP}(T) 
= \mathbb{E}\!\left[A_{\mathrm{Das}}(T,\theta^\star)\right]
= C_D \, \mathbb{E}[L(\theta^\star)] 
  \left( 
    \sqrt{\frac{\log(p+1)}{B'}} + \frac{p}{B'} 
  \right).
\]
When $B' \gg p$, this term is of order $O(L(\theta^\star)\sqrt{\log p / B'})$ and is negligible relative to the learning regret of Theorem~\ref{thm:regret}.

To obtain a bound against the LP benchmark, let $\mathrm{OPT}_{LP}(\theta^\star)$ and $\mathrm{OPT}_{IP}(\theta^\star)$ denote the values of the LP relaxation and integer program under true means $\theta^\star$ respectively (Appendix~\ref{appx:lp}); by Lemma~3 of \citet{das2022budgeted}, $\mathrm{OPT}_{LP}(\theta^\star) \geq \mathrm{OPT}_{IP}(\theta^\star)$, so benchmarking against $\mathrm{OPT}_{LP}$ is conservative. 
Let $R_{\theta^\star}(\pi) = \mathbb{E}_{\pi}\!\left[\sum_{t=1}^T r_{\theta^\star}(S_t)\right]$ denote the expected total reward of policy $\pi$ under true means $\theta^\star$, and let $\pi^{LP}(\theta^\star)$ denote the CBwK-LP primal-dual policy with $\theta^\star$ supplied in place of posterior samples. 
Decomposing the total gap,
\begin{align*}
\mathrm{OPT}_{LP}(\theta^\star) - R_{\theta^\star}(\pi^{TS})
&= \underbrace{\left\{
    \mathrm{OPT}_{LP}(\theta^\star) 
    - R_{\theta^\star}(\pi^{LP}(\theta^\star))
  \right\}}_{\text{planning gap}} 
+ \underbrace{\left\{
    R_{\theta^\star}(\pi^{LP}(\theta^\star)) 
    - R_{\theta^\star}(\pi^{TS})
  \right\}}_{\text{learning gap}},
\end{align*}
the first term is controlled by the known-means analysis of \citet{das2022budgeted}; taking expectations over the prior on $\theta^\star$ gives $\mathrm{Gap}_{LP}(T) = \mathbb{E}[A_{\mathrm{Das}}(T,\theta^\star)]$, 
where $A_{\mathrm{Das}}$ is defined in Appendix~\ref{appx:lp}. 
The second term is bounded by Theorem~\ref{thm:regret}, since $\text{BayesRisk}(T,\pi^{TS}) = \mathbb{E}[\mathrm{OPT}_{LP}(\theta^\star) - R_{\theta^\star}(\pi^{TS})]$ controls the learning gap after taking expectations over the prior.

\begin{corollary}[LP-benchmark Bayes risk of CBwK-LP-TS]
\label{cor:lp}
Under the conditions of Theorem~\ref{thm:regret},
\begin{align*}
\mathrm{BayesRisk}_{LP}(T) 
&:= \mathbb{E}\!\left[
    \mathrm{OPT}_{LP}(\theta^\star) - R_{\theta^\star}(\pi^{TS})
   \right] \\
&\leq \mathrm{Gap}_{LP}(T) 
  + O\!\left(
      \sqrt{p\log(pT)
      \cdot\min\!\left(KT,\,\tfrac{B}{c_{\min}}\right)}
    \right).
\end{align*}
Since $\mathrm{OPT}_{IP}(\theta^\star)\leq\mathrm{OPT}_{LP}(\theta^\star)$, the same bound holds against the integer optimum.
\end{corollary}

\subsection{Fractional Rewards}
\label{appx:frac_rew}
\begin{corollary}[Bayes Risk under Fractional Rewards]
\label{cor:fractional}
Under fractional rewards and the pseudo-posterior Beta update 
\eqref{eq:beta_update},
\[
\mathrm{BayesRisk}(T, \pi^{TS}) \leq 
  O\!\left(\sqrt{p \log(pT) \cdot \min\!\left(KT,\, 
  \tfrac{B}{c_{\min}}\right)}\right) + 
  K\sum_{t=1}^T \mathbb{E}[\varepsilon_t],
\] where $\varepsilon_t = d_{\mathrm{TV}}(\tilde\Pi_t, \Pi_t \mid \mathcal{D}_{t-1})$ is the total variation distance between the pseudo-posterior and the true posterior at round $t$. The additional term is small when rewards concentrate near $0$ or $1$, in which case the fractional algorithm achieves the same asymptotic rate as the binary case.
\end{corollary}

The additional term $K\sum_{t=1}^T \mathbb{E}[\varepsilon_t] $ captures the price of using a pseudo-posterior. This term is small when the fractional rewards concentrate near $0 $ or $1 $ --- that is, when covariates are either clearly predictive or clearly not --- since in this regime the fractional and binary updates agree approximately and the pseudo-posterior is close to the true posterior. In high-dimensional settings where the signal is sparse, this condition is empirically plausible. 

\subsection{Proof of theorem \ref{thm:lower_bound}}
\label{appx:thm_lower}
Consider an instance where $c_j = c_{\min} $ for all $j \in [p] $. With uniform costs, the bang-per-buck ratio $\theta_j / c_j = \theta_j / c_{\min} $ is strictly monotone in $\theta_j $. Identification of the optimal super-arm must come entirely from reward observations.

Assign each arm a Bernoulli reward distribution with mean $\theta_j^* \geq 1/2 > 0 $. Because every arm has strictly 
positive expected reward, pulling an additional arm always increases expected reward, so any algorithm pulling fewer than $K $ arms in any round incurs strictly additional regret relative to the optimal policy. Without loss of generality we therefore restrict attention to algorithms that pull exactly $K $ arms per round on this instance. The condition $K < p $ ensures there exist suboptimal arms, making the identification problem nontrivial.

Because each arm pull costs exactly $c_{\min} $ regardless of which arm is selected, the total budget $B $ induces a deterministic stopping time: the experiment terminates after exactly $\lfloor B / c_{\min} \rfloor $ total arm pulls 
independently of the algorithm's choices. Combined with the horizon constraint of $KT $ total pulls, the effective total pull budget is:
\[
Q = \min\left(KT,\left\lfloor \tfrac{B}{c_{\min}} \right\rfloor\right).
\]
Since algorithms pull exactly $K $ arms per round on this instance, $Q $ total pulls correspond to $t^* = \lfloor Q / K \rfloor $ full rounds. A lower bound on regret over $t^* $ rounds is a valid lower bound over the full horizon, since rounds beyond $t^* $ are either infeasible due to budget exhaustion or contribute non-negatively to cumulative regret.

We apply the lower bound of \citet{kveton2015tight}, whose Proposition 2 is established for the case $K \mid p $ to construct a disjoint partition of arms into non-overlapping super-arms, ensuring independence between pulls. For general $p $, we restrict to a sub-instance of $p' = \lfloor p/K \rfloor \cdot K $ arms, which satisfies $K \mid p' $ and recovers the same asymptotic bound since $p' \geq p/2 $. 

The problem now reduces to a stochastic combinatorial semi-bandit with $p' $ base arms, super-arm size $K $, linear reward $r_{\theta^*}(S) = \sum_{j \in S} \theta_j^* $, rewards bounded in $[0,1] $, semi-bandit feedback over $t^* $ rounds. Proposition 2 of \citet{kveton2015tight}, which holds for any horizon $t>0$, proves a gap-free lower bound of $\Omega(\sqrt{Kp't^*}) $, under the only non-trivial setting of $Q\geq p$ where $B\geq p\cdot c_{\min}$ by assumption, and $T\geq p/K$ so that we can visit every arm. We get
\[
\mathbb{E}[\mathrm{Regret}(T,\pi)] \geq \Omega\left(\sqrt{K \cdot p' \cdot \tfrac{Q}{K}}\right) = \Omega(\sqrt{p'Q}) = \Omega(\sqrt{pQ}),
\]
where the last equality holds since $p' = \lfloor p/K \rfloor \cdot K \geq p/2 $, so $\sqrt{p'} = \Omega(\sqrt{p}) $. Substituting $Q = \min(KT, \lfloor B/c_{\min} \rfloor) $ yields the final bound:
\[
\mathbb{E}[\mathrm{Regret}(T,\pi)] \geq\Omega\left(\sqrt{p \cdot \min\left(KT, \tfrac{B}{c_{\min}}\right)}\right).
\]
This lower bound isolates the statistical learning complexity under uniform costs. Instances with severe cost heterogeneity may be strictly harder, potentially yielding a tighter lower bound than established here. Characterizing whether cost heterogeneity introduces an additional dominant complexity term remains an open question.

\section{Simulations}
\label{appx:sims}
\subsection{Data-Generating Processes}
\label{appx:dgp}

\textbf{Linear DGP.}
Covariates are drawn independently as $X \sim \mathcal{N}(0, I_p)$. Outcomes are
\[ Y_i(0) = X_i^\top\beta + 2X_{i,1}X_{i,5} + \varepsilon_i, \quad
\varepsilon_i \sim \mathcal{N}(0,1),
\]
with $\beta_j \sim \mathcal{N}(2, 0.1)$ for $j \leq 10$, $\beta_j \sim \mathcal{N}(-2, 0.05)$ for $j \in \{11,\ldots,20\}$, and $\beta_j \sim \mathcal{N}(0, 0.01)$ for $j > 20$. The weak but nonzero coefficients for $j > 20$ reflect a realistic setting where noise covariates carry negligible predictive signal rather than being exactly inert. This is intentional: DARTS need not recover the exact support of $\beta$, but rather concentrate budget on the covariates that most reduce $\mathrm{Var}(\hat\tau_t)$ --- a strictly easier and more practically relevant objective.

\textbf{Liang DGP.}
Each covariate is generated as $X_{ij} = (e_i + z_{ij})/2$ where $e_i, z_{ij} \overset{iid}{\sim} \mathcal{N}(0,1)$, then scaled to $[0,1]$ and mean-centered. Outcomes follow
\[ Y_i(0) = \tfrac{10X_{i,2}}{1+X_{i,1}^2} + 5\sin(X_{i,3}-X_{i,4}) + 2X_{i,5} + X_i^\top\beta + \varepsilon_i, \]
with coefficients as in the linear DGP.

\textbf{Treatment and ATE.} In both DGPs, $Y_i(1) = Y_i(0) + \tau$ with $\tau = 4$.

\subsection{Simulation Regimes}
\label{appx:regimes}
\textbf{Monte Carlo grid (Section~\ref{sec:sims_results}).} We use $T = 100$ batches of $n = \in \{100, 1000\}$ units, $p \in \{100, 1000\}$, budgets $B \in \{1000, 2000,  20,000\}$, and equal costs $c_j = 1$ for all $j$. Both DGPs are evaluated. Results are averaged over 1000 replications with seeds \texttt{set.seed(s)} for $s = 1, \ldots, 1000$.

\textbf{Method comparison (Section~\ref{sec:madcovar_armm}).} We use $T = 200$ batches of $n = 1000$ units, $p = 100$, and budget $B = 2000$. Variable costs $c_j \overset{iid}{\sim} \mathrm{Uniform}(0, 2)$ and the covariate set for ARMM and MADCovar are redrawn in each replication using the replication seed \texttt{set.seed(s)} for $s = 1, \ldots, 1000$, which also governs the data and assignment randomness; on average, approximately 3 of the 12 selected covariates overlap with the 
true oracle set.

\subsection{Hyperparameters and Implementation}
\label{appx:hyper}

Beta priors are initialized at $\alpha_j(0) = \beta_j(0) = 1$.
The shadow-price learning rate is $\varepsilon = \sqrt{\log(p+1)/B_0}$ where $B_0 = \min(B, T)$. Rerandomization uses $K = 5{,}000$ candidate assignments per batch; the accepted assignment is chosen by a fair coin flip between the minimizer and its complement. The LASSO penalty is selected by 5-fold cross-validation with the \texttt{lambda.min} rule on control units. 
If LASSO selects no covariate, all rewards for that batch are set to zero and the prior is unchanged. Covariate matrices with near-singular sample covariance are handled via the Moore--Penrose pseudoinverse (\texttt{ginv} in R). 
ARMM uses $\rho = 0.75$ and ridge stabilization $\lambda = 10^{-4}$.
MADCovar uses plain OLS nuisance models with $p_{\mathrm{mad}} = 0.5$.

\subsection{Compute Resources}
\label{appx:resources}
Single-run analyses and diagnostic plots were conducted locally on an Apple M1 processor (8 cores) running R version 4.5.0. Monte Carlo replications were parallelized across independent jobs on an institutional high-performance computing cluster using bash job submission scripts, with each replication assigned an independent random seed. 

For the primary method comparison ($n = 1000$, $T = 200$, $p = 100$, $K = 5{,}000$ rerandomization candidates), cluster jobs were allocated 1 CPU core each. The algorithm reached an observed peak memory consumption of approximately 1.4GB. 
Each replication required approximately 35 CPU minutes.

To optimize cluster throughput for the extensive Monte Carlo grid, the 1000-replication studies were distributed as an array of 100 jobs, with each job sequentially executing 10 replications on a single CPU core. The heaviest grid settings ($n = 1000$) peaked at approximately 1.1GB of RAM per job. 
The smaller grid settings ($n = 100$, $T = 100$) required substantially less compute time and had a peak memory footprint of only 250MB. 
All code was implemented in R using the \texttt{tidyverse}, \texttt{glmnet}, \texttt{car}, and \texttt{MASS} packages.

\subsection{Inference}
\label{appx:inference}

Batch-level ATE estimates use Lin's regression adjustment with HC2 heteroskedasticity-robust standard errors. The cumulative estimate $\hat{\mu}_T$ is the inverse-variance weighted combination of batch-level estimates as in equation~\eqref{eq:meta}. All reported confidence intervals are fixed-$T$ Wald intervals at the 95\% level, $\hat{\mu}_T \pm 1.96\,\hat{\sigma}_T$.
\subsection{Robustness Checks}
\label{appx:robustness}

\subsubsection{Heterogeneous Treatment Effects}

We modify the method comparison DGP to introduce individual-level treatment effect heterogeneity. The outcome surface for $Y_i(0)$ follows the Liang structure on covariates 1--20, while 
the individual treatment effect is
\[
\tau_i = 2 + 3\sin(\pi X_{i,6}X_{i,7}) + 4(X_{i,8}-0.5)^2 + 
2X_{i,9} + X_{i,10},
\]
a Friedman-style function of covariates 6--10. The true ATE is $\tau = \mathbb{E}[\tau_i] \approx 5.6451$, with substantial individual-level variation (between 2.64 and 8.28).  All other design parameters match the method comparison ($T = 200$, $n = 1000$, $p = 100$, $B = 2000$, variable costs,\texttt{set.seed(7)}).

This setup tests whether DARTS, which uses a reward signal based on $Y_i(0,t)$ predictiveness among control units, still concentrates budget on useful covariates when some oracle variables are effect modifiers rather than purely prognostic. Results are presented as a single illustrative run. 

\subsubsection{Oracle-Costly Covariates}
\begin{figure}[ht]
  \centering
  \begin{subfigure}[t]{0.46\textwidth}
    \centering
    \includegraphics[width=\textwidth]{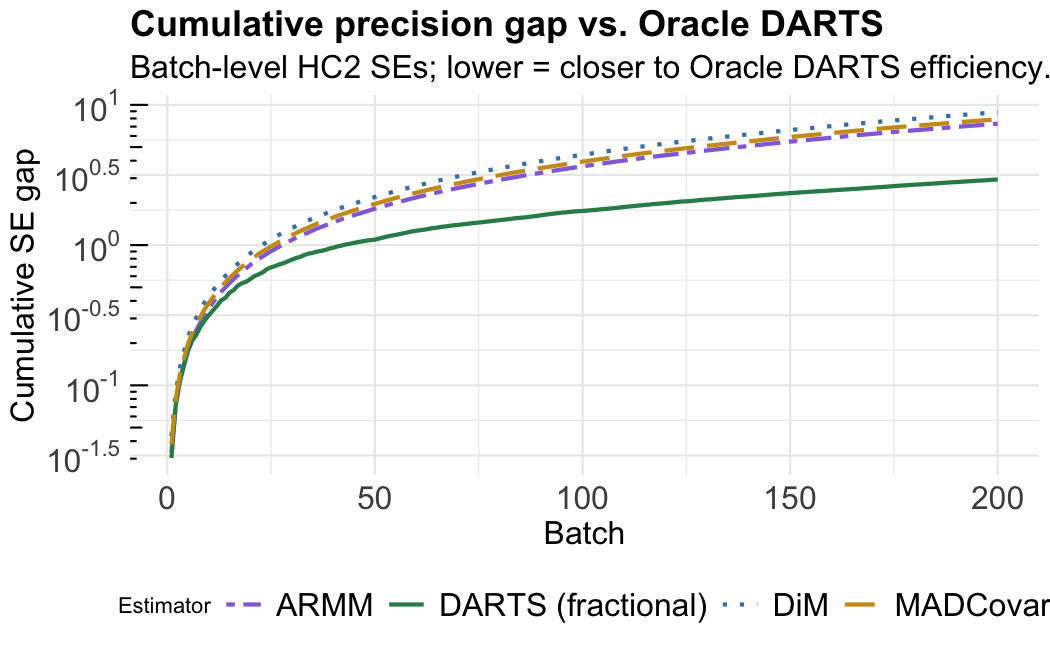}
    \caption{Cumulative regret relative to Oracle DARTS. Single run, 
    \texttt{set.seed(7)}.}
  \end{subfigure}
  \hfill
  \begin{subfigure}[t]{0.46\textwidth}
    \centering
    \includegraphics[width=\textwidth]{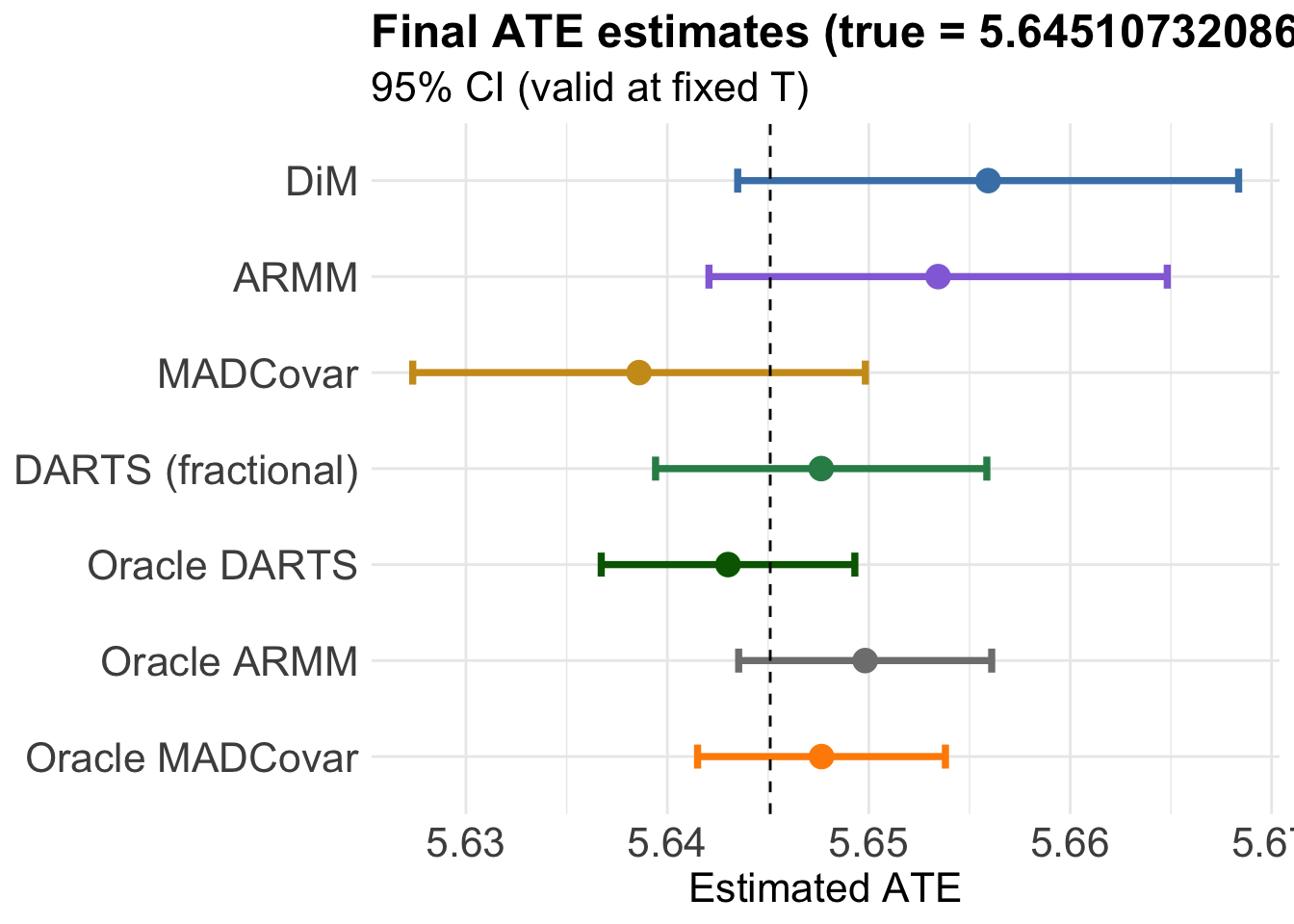}
    \caption{Final ATE estimates with 95\% CIs at $T = 200$.}
  \end{subfigure}
  \caption{Heterogeneous treatment effects robustness check. Liang outcome surface on covariates 1--20 with Friedman-style CATE on covariates 6--10; true ATE $\approx 5.645$. Single run ($n = 1000$,  $T = 200$, $p = 100$, $B = 2000$, variable costs). ARMM and MADCovar receive randomly pre-selected covariates; DARTS learns 
  from the full pool. Oracle variants use the true 20 signal  covariates. All CIs are 95\% valid at fixed $T$; HC2 standard errors throughout.}
  \label{fig:robustness_hetero}
\end{figure}

\begin{figure}[ht]
  \centering
  \begin{subfigure}[t]{0.46\textwidth}
    \centering
    \includegraphics[width=\textwidth]{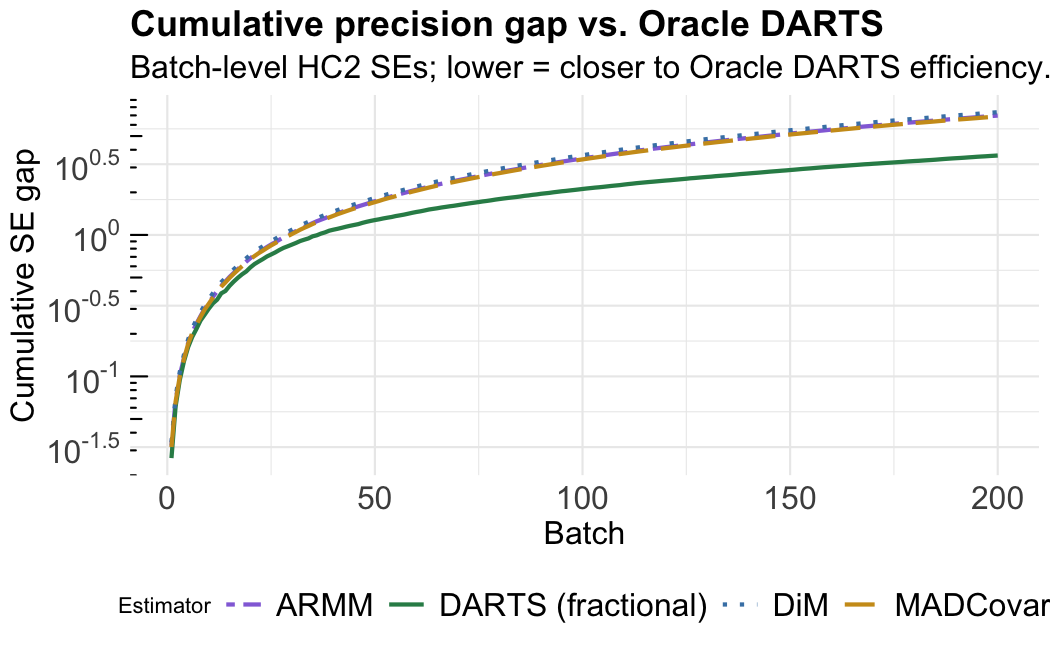}
    \caption{Cumulative regret relative to Oracle DARTS. Single run, \texttt{set.seed(7)}.}
  \end{subfigure}
  \hfill
  \begin{subfigure}[t]{0.46\textwidth}
    \centering
    \includegraphics[width=\textwidth]{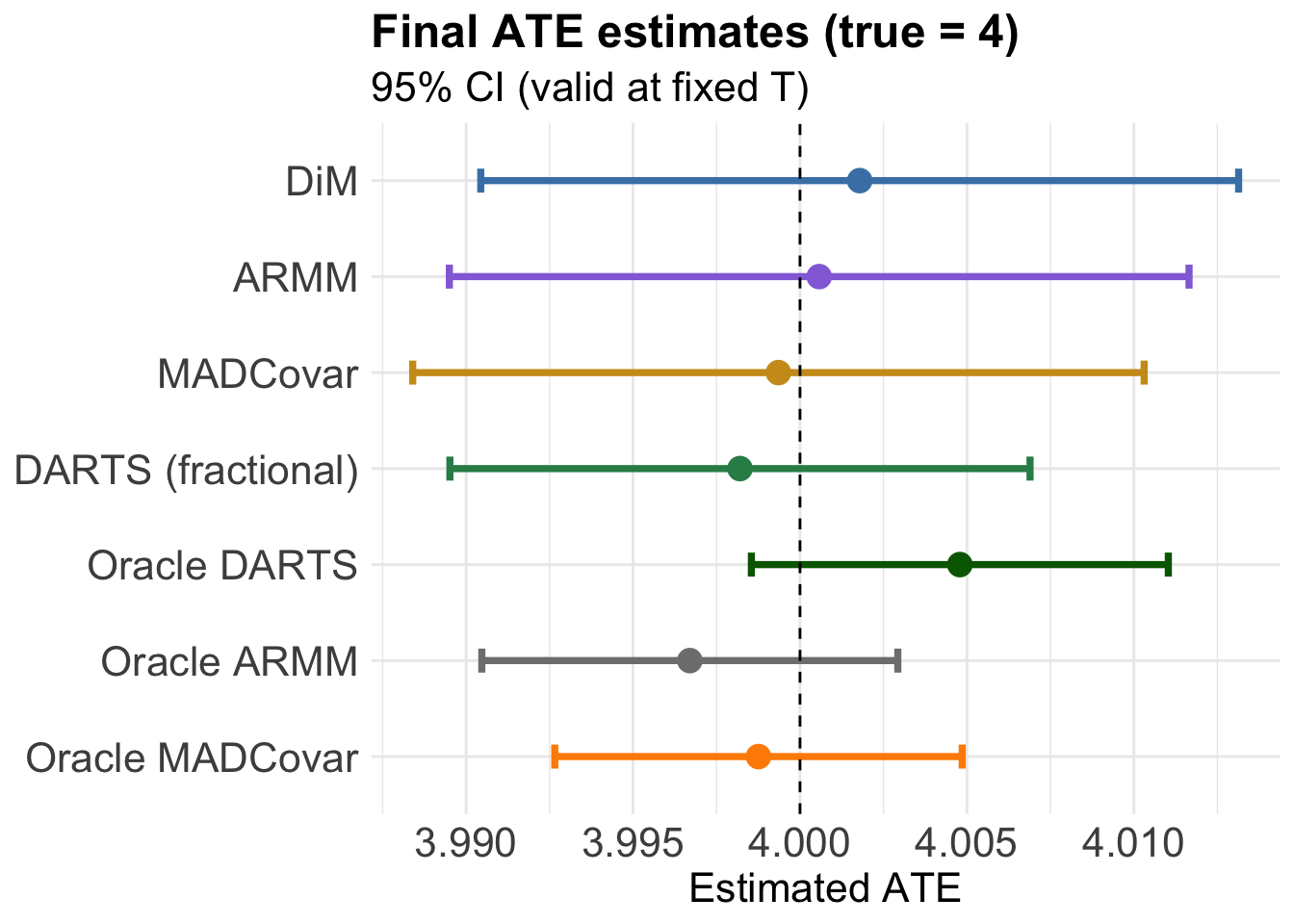}
    \caption{Final ATE estimates with 95\% CIs at $T = 200$.}
  \end{subfigure}
  \caption{Oracle-costly covariates robustness check. Liang DGP with  $\tau = 4$; oracle covariates ($j \leq 20$) assigned cost $c_j = 1.1$ and noise covariates cost $c_j = 0.8$, rescaled by their maximum with $B = 2000$ rescaled accordingly. Single run ($n = 1000$, $T = 200$, $p = 100$). ARMM and MADCovar receive randomly pre-selected covariates; DARTS learns from the full pool.  Oracle variants use the true 20 signal covariates. All CIs are 95\% valid at fixed $T$; HC2 standard errors throughout.}
  \label{fig:robustness_costly}
\end{figure}

We use the Liang DGP with $\tau = 4$ but assign differential costs: oracle covariates ($j \leq 20$) have cost $c_j = 1.1$ and noise covariates ($j > 20$) have cost $c_j = 0.8$, with $B = 2000$. This tests whether DARTS still concentrates budget on prognostic covariates when doing so is relatively more expensive. All other design parameters match the method comparison (\texttt{set.seed(7)}).
In both robustness checks (Figures ~\ref{fig:robustness_hetero},\ref{fig:robustness_costly}), DARTS closes a substantial share of the efficiency gap to Oracle DARTS, consistent with the main method comparison results.

\subsection{Diagnostic Plots}
\label{appx:diagnostics}
\begin{figure}[ht]
  \centering
  \begin{subfigure}[t]{0.46\textwidth}
    \centering
    \includegraphics[width=\textwidth]{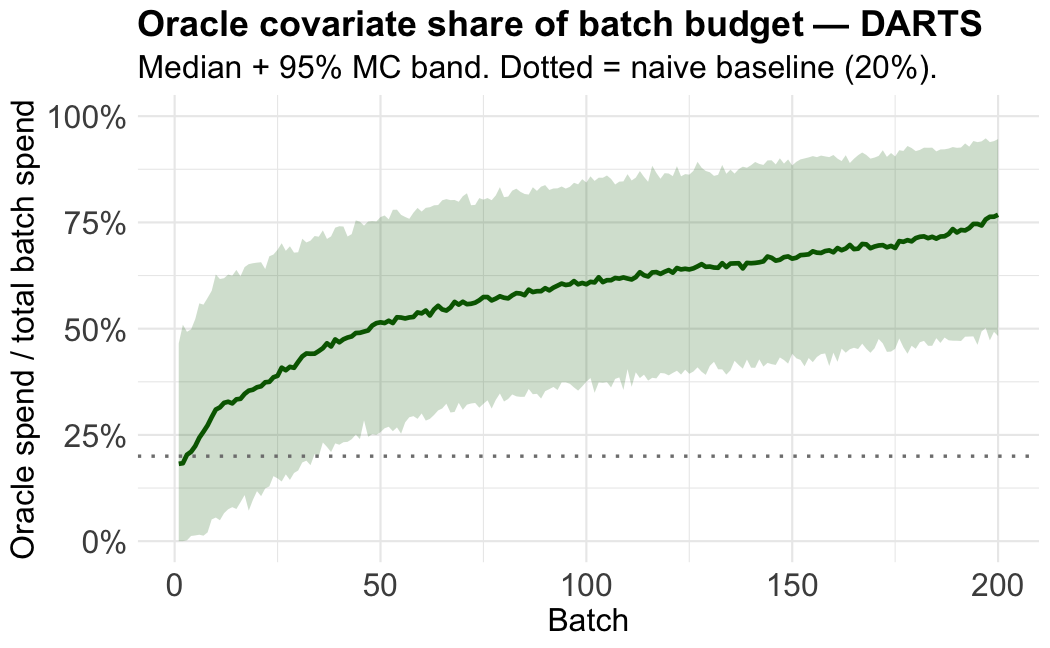}
    \caption{Median share of batch budget allocated to oracle 
    covariates (covariates 1--20) across 1000 replications. Solid 
    line shows median; shaded band shows 95\% interval from run 
    percentiles.}
    \label{fig:oracle_budget}
  \end{subfigure}
  \hfill
  \begin{subfigure}[t]{0.46\textwidth}
    \centering
    \includegraphics[width=\textwidth]{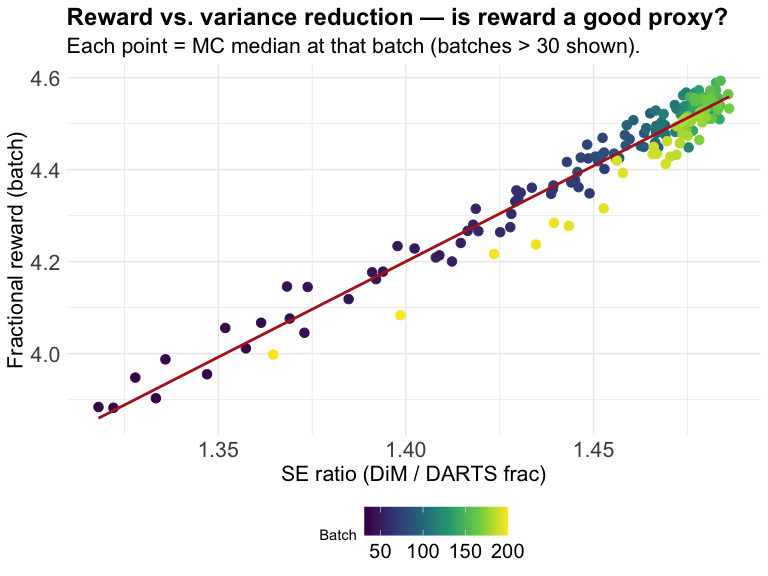}
    \caption{Batch-level LASSO fractional reward against SE ratio 
    $\hat{\sigma}_{\mathrm{DiM},t} / \hat{\sigma}_{\mathrm{DARTS},t}$
    after a 30-batch burn-in, pooled across 1000 replications. Each 
    point is one batch from one replication. Correlation coefficient is 0.977.}
    \label{fig:reward_se}
  \end{subfigure}
  \caption{Diagnostics from the 1000-replication method comparison 
  (Liang DGP, $n = 1000$, $T = 200$, $p = 100$, $B = 2000$, 
  variable costs). Left: the oracle budget share increases 
  monotonically, confirming progressive concentration of resources 
  on prognostic covariates. Right: the positive reward--SE 
  correlation confirms that the LASSO fractional reward proxy 
  translates learning into precision gains.}
  \label{fig:diagnostics}
\end{figure}

\textbf{Oracle budget share.} Figure~\ref{fig:oracle_budget} shows the median share of batch budget allocated to oracle covariates (covariates 1--20) across 1000 replications of the method comparison, with 95\% bands from run percentiles. The share increases monotonically from near-random in early batches toward a stable plateau, confirming progressive concentration of resources on prognostic covariates.

\textbf{Reward--SE correlation.} Figure~\ref{fig:reward_se} plots batch-level LASSO fractional rewards against the SE ratio $\hat{\sigma}_{\mathrm{DiM},t} / \hat{\sigma}_{\mathrm{DARTS},t}$ for each batch after a 30-batch burn-in, pooled across 1000 replications. The positive correlation confirms that batches in which DARTS selects more prognostic covariates --- as measured by the reward signal --- yield larger reductions in batch-level standard error relative to DiM.

\begin{longtable}{@{} l l ccccc @{}}
\caption{%
  Rerandomisation vs.\ rerandomisation with adjustment in DARTS designs.
  \underline{DiM} and \underline{Oracle} are benchmark rows; Oracle is
  separated by a partial rule within each setup block.
  Bold marks the best non-benchmark value within each setup.
  RE vs.\ DiM is the ratio of DiM MSE to the method's MSE,
  estimated across $1{,}000$ simulation replications;
  \% Red.\ is the percentage reduction in MSE relative to DiM.
  $B$ = total budget, $p$ = covariates, $n$ = units/batch, $T$ = batches.}
\label{tab:rerand-vs-adj} \\

\toprule
\textbf{Setup} & \textbf{Method}
  & \textbf{Emp SD} & \textbf{Bias} & \textbf{MSE}
  & \textbf{RE vs DiM} & \textbf{\% Red.} \\
\midrule
\endfirsthead

\multicolumn{7}{l}{\textit{Table \ref{tab:rerand-vs-adj} continued from previous page.}} \\[4pt]
\toprule
\textbf{Setup} & \textbf{Method}
  & \textbf{Emp SD} & \textbf{Bias} & \textbf{MSE}
  & \textbf{RE vs DiM} & \textbf{\% Red.} \\
\midrule
\endhead

\midrule
\multicolumn{7}{r}{\textit{Continued on next page.}} \\
\endfoot

\bottomrule
\endlastfoot

\multicolumn{7}{@{}l}{\textit{LIANG \;|\; $B{=}1000,\; p{=}100,\; n{=}100,\; T{=}100$}} \\[2pt]
 & \underline{DiM}         & 0.0302 & $3{\times}10^{-4}$  & $9{\times}10^{-4}$ & 1.0000          & 0.00           \\[1pt]
 & Rerand (DARTS)          & 0.0253 & $-9{\times}10^{-4}$ & $6{\times}10^{-4}$ & 1.4216          & 29.65          \\[1pt]
 & Rerand+Adj (DARTS)      & 0.0245 & $-1{\times}10^{-3}$ & $6{\times}10^{-4}$ & \textbf{1.5251} & \textbf{34.43} \\[1pt]
\cmidrule(l){2-7}
 & \underline{Oracle}      & 0.0153 & $1{\times}10^{-3}$  & $2{\times}10^{-4}$ & 3.8874          & 74.28          \\
\midrule

\multicolumn{7}{@{}l}{\textit{LIANG \;|\; $B{=}2000,\; p{=}100,\; n{=}100,\; T{=}100$}} \\[2pt]
 & \underline{DiM}         & 0.0308 & $-4{\times}10^{-4}$ & $9{\times}10^{-4}$ & 1.0000          & 0.00           \\[1pt]
 & Rerand (DARTS)          & 0.0249 & $-1{\times}10^{-3}$ & $6{\times}10^{-4}$ & 1.5355          & 34.88          \\[1pt]
 & Rerand+Adj (DARTS)      & 0.0218 & $-6{\times}10^{-4}$ & $5{\times}10^{-4}$ & \textbf{1.9918} & \textbf{49.79} \\[1pt]
\cmidrule(l){2-7}
 & \underline{Oracle}      & 0.0153 & $1{\times}10^{-3}$  & $2{\times}10^{-4}$ & 4.0421          & 75.26          \\
\midrule

\multicolumn{7}{@{}l}{\textit{LIANG \;|\; $B{=}2000,\; p{=}100,\; n{=}1000,\; T{=}100$}} \\[2pt]
 & \underline{DiM}         & 0.0086 & $-5{\times}10^{-4}$ & $1{\times}10^{-4}$ & 1.0000          & 0.00           \\[1pt]
 & Rerand (DARTS)          & 0.0064 & $-1{\times}10^{-4}$ & $0$                & 1.7856          & 43.99          \\[1pt]
 & Rerand+Adj (DARTS)      & 0.0057 & $0$                 & $0$                & \textbf{2.3177} & \textbf{56.85} \\[1pt]
\cmidrule(l){2-7}
 & \underline{Oracle}      & 0.0045 & $1{\times}10^{-4}$  & $0$                & 3.6377          & 72.51          \\
\midrule

\multicolumn{7}{@{}l}{\textit{LIANG \;|\; $B{=}2000,\; p{=}1000,\; n{=}100,\; T{=}100$}} \\[2pt]
 & \underline{DiM}         & 0.0296 & $-0.0003$ & $9{\times}10^{-4}$ & 1.0000          &  0.00          \\[1pt]
 & Rerand (DARTS)          & 0.0294 &  $0.0004$ & $9{\times}10^{-4}$ & \textbf{1.0144} & \textbf{1.42}  \\[1pt]
 & Rerand+Adj (DARTS)      & 0.0302 &  $0.0002$ & $9{\times}10^{-4}$ & 0.9573          & $-4.46$        \\[1pt]
\cmidrule(l){2-7}
 & \underline{Oracle}      & 0.0153 & $-0.0012$ & $2{\times}10^{-4}$ & 3.7321          & 73.21          \\
\midrule

\multicolumn{7}{@{}l}{\textit{LIANG \;|\; $B{=}2000,\; p{=}1000,\; n{=}1000,\; T{=}100$}} \\[2pt]
 & \underline{DiM}         & 0.0083 & $4{\times}10^{-4}$  & $1{\times}10^{-4}$ & 1.0000          & 0.00           \\[1pt]
 & Rerand (DARTS)          & 0.0078 & $-1{\times}10^{-4}$ & $1{\times}10^{-4}$ & 1.1470          & 12.81          \\[1pt]
 & Rerand+Adj (DARTS)      & 0.0075 & $0$                 & $1{\times}10^{-4}$ & \textbf{1.2309} & \textbf{18.76} \\[1pt]
\cmidrule(l){2-7}
 & \underline{Oracle}      & 0.0045 & $0$                 & $0$                & 3.4878          & 71.32          \\
\midrule

\multicolumn{7}{@{}l}{\textit{LIANG \;|\; $B{=}20000,\; p{=}100,\; n{=}1000,\; T{=}100$}} \\[2pt]
 & \underline{DiM}         & 0.0089 & $-2{\times}10^{-4}$ & $1{\times}10^{-4}$ & 1.0000          & 0.00           \\[1pt]
 & Rerand (DARTS)          & 0.0071 & $-2{\times}10^{-4}$ & $0$                & 1.5726          & 36.41          \\[1pt]
 & Rerand+Adj (DARTS)      & 0.0049 & $0$                 & $0$                & \textbf{3.2967} & \textbf{69.67} \\[1pt]
\cmidrule(l){2-7}
 & \underline{Oracle}      & 0.0045 & $1{\times}10^{-4}$  & $0$                & 3.9257          & 74.53          \\
\midrule\midrule

\multicolumn{7}{@{}l}{\textit{LINEAR \;|\; $B{=}1000,\; p{=}100,\; n{=}100,\; T{=}100$}} \\[2pt]
 & \underline{DiM}         & 0.1835 & 0.0043   & 0.0337 &  1.0000          &  0.00          \\[1pt]
 & Rerand (DARTS)          & 0.1698 & 0.0024   & 0.0288 &  \textbf{1.1681} & \textbf{14.39} \\[1pt]
 & Rerand+Adj (DARTS)      & 0.1704 & 0.0026   & 0.0290 &  1.1606          & 13.84          \\[1pt]
\cmidrule(l){2-7}
 & \underline{Oracle}      & 0.0465 & 0.0012   & 0.0022 & 15.5885          & 93.59          \\
\midrule

\multicolumn{7}{@{}l}{\textit{LINEAR \;|\; $B{=}2000,\; p{=}100,\; n{=}100,\; T{=}100$}} \\[2pt]
 & \underline{DiM}         & 0.1871 & $-0.0010$ & 0.0350 &  1.0000          &  0.00          \\[1pt]
 & Rerand (DARTS)          & 0.1514 &  $0.0087$ & 0.0230 &  1.5223          & 34.31          \\[1pt]
 & Rerand+Adj (DARTS)      & 0.1206 &  $0.0053$ & 0.0146 &  \textbf{2.3994} & \textbf{58.32} \\[1pt]
\cmidrule(l){2-7}
 & \underline{Oracle}      & 0.0449 & $-0.0007$ & 0.0020 & 17.3388          & 94.23          \\
\midrule

\multicolumn{7}{@{}l}{\textit{LINEAR \;|\; $B{=}2000,\; p{=}100,\; n{=}1000,\; T{=}100$}} \\[2pt]
 & \underline{DiM}         & 0.0581 & $-0.0014$ & 0.0034 &  1.0000          &  0.00          \\[1pt]
 & Rerand (DARTS)          & 0.0518 &  $0.0011$ & 0.0027 &  1.2605          & 20.55          \\[1pt]
 & Rerand+Adj (DARTS)      & 0.0308 &  $0.0007$ & 0.0009 &  \textbf{3.5633} & \textbf{71.93} \\[1pt]
\cmidrule(l){2-7}
 & \underline{Oracle}      & 0.0139 &  $0.0006$ & 0.0002 & 17.5896          & 94.31          \\
\midrule

\multicolumn{7}{@{}l}{\textit{LINEAR \;|\; $B{=}2000,\; p{=}1000,\; n{=}100,\; T{=}100$}} \\[2pt]
 & \underline{DiM}         & 0.1869 &  $0.0046$ & 0.0349 &  1.0000          &  0.00          \\[1pt]
 & Rerand (DARTS)          & 0.1837 & $-0.0006$ & 0.0337 &  \textbf{1.0358} & \textbf{3.46}  \\[1pt]
 & Rerand+Adj (DARTS)      & 0.1898 &  $0.0018$ & 0.0360 &  0.9701          & $-3.08$        \\[1pt]
\cmidrule(l){2-7}
 & \underline{Oracle}      & 0.0477 & $-0.0015$ & 0.0023 & 15.3760          & 93.50          \\
\midrule

\multicolumn{7}{@{}l}{\textit{LINEAR \;|\; $B{=}2000,\; p{=}1000,\; n{=}1000,\; T{=}100$}} \\[2pt]
 & \underline{DiM}         & 0.0588 & $-0.0059$ & 0.0035 &  1.0000          &  0.00          \\[1pt]
 & Rerand (DARTS)          & 0.0600 &  $0.0025$ & 0.0036 &  0.9677          & $-3.37$        \\[1pt]
 & Rerand+Adj (DARTS)      & 0.0529 &  $0.0007$ & 0.0028 &  \textbf{1.2480} & \textbf{19.84} \\[1pt]
\cmidrule(l){2-7}
 & \underline{Oracle}      & 0.0144 &  $0.0003$ & 0.0002 & 16.9663          & 94.10          \\

\end{longtable}


\begin{longtable}{@{} l l cccccc @{}}
\caption{%
  Performance of rerandomisation with adjustment (DARTS) versus
  rerandomisation with randomly selected covariates (Random),
  with \underline{DiM} and \underline{Oracle} benchmarks.
  Oracle is separated by a partial rule within each setup block.
  Bold marks the best non-benchmark RE vs.\ DiM within each setup.
  Median SE is the median estimated standard error across replications;
  Coverage is the empirical 95\% CI coverage rate;
  RE vs.\ DiM is the ratio of DiM MSE to the method's MSE,
  estimated across $1{,}000$ simulation replications.}
\label{tab:DARTS-vs-random} \\

\toprule
\textbf{Setup} & \textbf{Method}
  & \textbf{Bias} & \textbf{Emp SD} & \textbf{MSE}
  & \textbf{Median SE} & \textbf{Coverage} & \textbf{RE vs DiM} \\
\midrule
\endfirsthead

\multicolumn{8}{l}{\textit{Table \ref{tab:DARTS-vs-random} continued from previous page.}} \\[4pt]
\toprule
\textbf{Setup} & \textbf{Method}
  & \textbf{Bias} & \textbf{Emp SD} & \textbf{MSE}
  & \textbf{Median SE} & \textbf{Coverage} & \textbf{RE vs DiM} \\
\midrule
\endhead

\midrule
\multicolumn{8}{r}{\textit{Continued on next page.}} \\
\endfoot

\bottomrule
\endlastfoot

\multicolumn{8}{@{}l}{\textit{LIANG \;|\; $B{=}1000,\; p{=}100,\; n{=}100,\; T{=}100$}} \\[2pt]
 & \underline{DiM}    & $3{\times}10^{-4}$  & 0.0302 & $9{\times}10^{-4}$ & 0.0294 & 0.949 & 1.0000           \\[1pt]
 & Random             & $3{\times}10^{-4}$  & 0.0280 & $8{\times}10^{-4}$ & 0.0277 & 0.946 & 1.1623           \\[1pt]
 & DARTS              & $-1{\times}10^{-3}$ & 0.0245 & $6{\times}10^{-4}$ & 0.0231 & 0.943 & \textbf{1.5251}  \\[1pt]
\cmidrule(l){2-8}
 & \underline{Oracle} & $1{\times}10^{-3}$  & 0.0153 & $2{\times}10^{-4}$ & 0.0146 & 0.939 & 3.8874           \\
\midrule

\multicolumn{8}{@{}l}{\textit{LIANG \;|\; $B{=}2000,\; p{=}100,\; n{=}100,\; T{=}100$}} \\[2pt]
 & \underline{DiM}    & $-4{\times}10^{-4}$ & 0.0308 & $9{\times}10^{-4}$ & 0.0294 & 0.933 & 1.0000           \\[1pt]
 & Random             & $2{\times}10^{-4}$  & 0.0290 & $8{\times}10^{-4}$ & 0.0270 & 0.933 & 1.1323           \\[1pt]
 & DARTS              & $-6{\times}10^{-4}$ & 0.0218 & $5{\times}10^{-4}$ & 0.0208 & 0.939 & \textbf{1.9918}  \\[1pt]
\cmidrule(l){2-8}
 & \underline{Oracle} & $1{\times}10^{-3}$  & 0.0153 & $2{\times}10^{-4}$ & 0.0146 & 0.937 & 4.0421           \\
\midrule

\multicolumn{8}{@{}l}{\textit{LIANG \;|\; $B{=}2000,\; p{=}100,\; n{=}1000,\; T{=}100$}} \\[2pt]
 & \underline{DiM}    & $-5{\times}10^{-4}$ & 0.0086 & $1{\times}10^{-4}$ & 0.0085 & 0.946 & 1.0000           \\[1pt]
 & Random             & $1{\times}10^{-4}$  & 0.0075 & $1{\times}10^{-4}$ & 0.0077 & 0.960 & 1.3261           \\[1pt]
 & DARTS              & $0$                 & 0.0057 & $0$                & 0.0057 & 0.956 & \textbf{2.3177}  \\[1pt]
\cmidrule(l){2-8}
 & \underline{Oracle} & $1{\times}10^{-4}$  & 0.0045 & $0$                & 0.0045 & 0.948 & 3.6377           \\
\midrule

\multicolumn{8}{@{}l}{\textit{LIANG \;|\; $B{=}2000,\; p{=}1000,\; n{=}100,\; T{=}100$}} \\[2pt]
 & \underline{DiM}    & $-0.0003$ & 0.0296 & $9{\times}10^{-4}$ & 0.0294 & 0.945 & \textbf{1.0000}           \\[1pt]
 & Random             &  $0.0023$ & 0.0315 & $1{\times}10^{-3}$ & 0.0293 & 0.939 & 0.8810           \\[1pt]
 & DARTS              &  $0.0002$ & 0.0302 & $9{\times}10^{-4}$ & 0.0285 & 0.937 & 0.9573  \\[1pt]
\cmidrule(l){2-8}
 & \underline{Oracle} & $-0.0012$ & 0.0153 & $2{\times}10^{-4}$ & 0.0146 & 0.942 & 3.7321           \\
\midrule

\multicolumn{8}{@{}l}{\textit{LIANG \;|\; $B{=}2000,\; p{=}1000,\; n{=}1000,\; T{=}100$}} \\[2pt]
 & \underline{DiM}    & $4{\times}10^{-4}$  & 0.0083 & $1{\times}10^{-4}$ & 0.0085 & 0.959 & 1.0000           \\[1pt]
 & Random             & $1{\times}10^{-4}$  & 0.0085 & $1{\times}10^{-4}$ & 0.0083 & 0.944 & 0.9525           \\[1pt]
 & DARTS              & $0$                 & 0.0075 & $1{\times}10^{-4}$ & 0.0076 & 0.954 & \textbf{1.2309}  \\[1pt]
\cmidrule(l){2-8}
 & \underline{Oracle} & $0$                 & 0.0045 & $0$                & 0.0045 & 0.953 & 3.4878           \\
\midrule

\multicolumn{8}{@{}l}{\textit{LIANG \;|\; $B{=}20000,\; p{=}100,\; n{=}1000,\; T{=}100$}} \\[2pt]
 & \underline{DiM}    & $-2{\times}10^{-4}$ & 0.0089 & $1{\times}10^{-4}$ & 0.0085 & 0.948 & 1.0000           \\[1pt]
 & Random             & $-3{\times}10^{-4}$ & 0.0045 & $0$                & 0.0047 & 0.953 & \textbf{3.8763}  \\[1pt]
 & DARTS              & $0$                 & 0.0049 & $0$                & 0.0047 & 0.940 & 3.2967           \\[1pt]
\cmidrule(l){2-8}
 & \underline{Oracle} & $1{\times}10^{-4}$  & 0.0045 & $0$                & 0.0045 & 0.952 & 3.9257           \\
\midrule\midrule

\multicolumn{8}{@{}l}{\textit{LINEAR \;|\; $B{=}1000,\; p{=}100,\; n{=}100,\; T{=}100$}} \\[2pt]
 & \underline{DiM}    & 0.0043    & 0.1835 & 0.0337 & 0.1824 & 0.952 &  1.0000          \\[1pt]
 & Random             & $-0.0008$ & 0.1718 & 0.0295 & 0.1736 & 0.957 &  1.1413          \\[1pt]
 & DARTS              & 0.0026    & 0.1704 & 0.0290 & 0.1610 & 0.944 &  \textbf{1.1606} \\[1pt]
\cmidrule(l){2-8}
 & \underline{Oracle} & 0.0012    & 0.0465 & 0.0022 & 0.0442 & 0.934 & 15.5885          \\
\midrule

\multicolumn{8}{@{}l}{\textit{LINEAR \;|\; $B{=}2000,\; p{=}100,\; n{=}100,\; T{=}100$}} \\[2pt]
 & \underline{DiM}    & $-0.0010$ & 0.1871 & 0.0350 & 0.1824 & 0.944 &  1.0000          \\[1pt]
 & Random             & $-0.0095$ & 0.1726 & 0.0298 & 0.1677 & 0.945 &  1.1714          \\[1pt]
 & DARTS              &  $0.0053$ & 0.1206 & 0.0146 & 0.1142 & 0.937 &  \textbf{2.3994} \\[1pt]
\cmidrule(l){2-8}
 & \underline{Oracle} & $-0.0007$ & 0.0449 & 0.0020 & 0.0442 & 0.953 & 17.3388          \\
\midrule

\multicolumn{8}{@{}l}{\textit{LINEAR \;|\; $B{=}2000,\; p{=}100,\; n{=}1000,\; T{=}100$}} \\[2pt]
 & \underline{DiM}    & $-0.0014$ & 0.0581 & 0.0034 & 0.0583 & 0.953 &  1.0000          \\[1pt]
 & Random             &  $0.0011$ & 0.0518 & 0.0027 & 0.0583 & 0.977 &  1.2605          \\[1pt]
 & DARTS              &  $0.0007$ & 0.0308 & 0.0009 & 0.0583 & 0.998 &  \textbf{3.5633} \\[1pt]
\cmidrule(l){2-8}
 & \underline{Oracle} &  $0.0006$ & 0.0139 & 0.0002 & 0.0142 & 0.956 & 17.5896          \\
\midrule

\multicolumn{8}{@{}l}{\textit{LINEAR \;|\; $B{=}2000,\; p{=}1000,\; n{=}100,\; T{=}100$}} \\[2pt]
 & \underline{DiM}    &  $0.0046$ & 0.1869 & 0.0349 & 0.1825 & 0.944 &  \textbf{1.0000}          \\[1pt]
 & Random             &  $0.0019$ & 0.1895 & 0.0359 & 0.1852 & 0.953 &  0.9736          \\[1pt]
 & DARTS              &  $0.0018$ & 0.1898 & 0.0360 & 0.1846 & 0.943 &  0.9701 \\[1pt]
\cmidrule(l){2-8}
 & \underline{Oracle} & $-0.0015$ & 0.0477 & 0.0023 & 0.0446 & 0.931 & 15.3760          \\
\midrule

\multicolumn{8}{@{}l}{\textit{LINEAR \;|\; $B{=}2000,\; p{=}1000,\; n{=}1000,\; T{=}100$}} \\[2pt]
 & \underline{DiM}    & $-0.0059$ & 0.0588 & 0.0035 & 0.0583 & 0.946 &  1.0000          \\[1pt]
 & Random             &  $0.0025$ & 0.0600 & 0.0036 & 0.0583 & 0.939 &  0.9677          \\[1pt]
 & DARTS              &  $0.0007$ & 0.0529 & 0.0028 & 0.0583 & 0.964 &  \textbf{1.2480} \\[1pt]
\cmidrule(l){2-8}
 & \underline{Oracle} &  $0.0003$ & 0.0144 & 0.0002 & 0.0143 & 0.953 & 16.9663          \\

\end{longtable}

\begin{table}[ht]
\centering
\caption{%
  Performance at $T{=}200$ batches of $n{=}1000$ units, $p{=}100$ 
  covariates, $B{=}2000$, variable costs $c_j \overset{iid}{\sim} 
  \mathrm{Uniform}(0,2)$. True ATE $= 4$. Summary across 1000 MC 
  replications. \underline{DiM} is the baseline benchmark; oracle 
  methods use the true 20 signal covariates and are separated by a 
  partial rule. Bold marks the best non-benchmark, non-oracle value. 
  Rel.\ RMSE $= \mathrm{RMSE}_{\mathrm{DiM}} / 
  \mathrm{RMSE}_{\mathrm{method}}$; values ${>}1$ indicate improvement 
  over DiM. CI Width $= 2 \times 1.96 \times \text{Med.\ SE}$. 
  All CIs are 95\% (valid at fixed $T$).}
\label{tab:dgp1-performance-mc}
\small
\setlength{\tabcolsep}{5pt}
\begin{tabular}{@{} l cccccc @{}}
\toprule
\textbf{Method}
  & \textbf{Mean ATE} & \textbf{Bias} & \textbf{RMSE}
  & \textbf{Med.\ SE} & \textbf{CI Width} & \textbf{Rel.\ RMSE} \\
\midrule
 \underline{DiM}      & 4.0003 & $+3{\times}10^{-4}$ & 0.0058 & 0.0059 & 0.0230 & 1.000 \\[1pt]
 ARMM                 & 4.0000 & $0$                 & 0.0057 & 0.0056 & 0.0221 & 1.018 \\[1pt]
 MADCovar             & 4.0000 & $0$                 & 0.0056 & 0.0056 & 0.0219 & 1.036 \\[1pt]
 DARTS                & 3.9998 & $-2{\times}10^{-4}$ & \textbf{0.0041} & \textbf{0.0041} & 0.0162 & \textbf{1.415} \\[1pt]
\cmidrule(l){1-7}
 \underline{Oracle DARTS}    & 3.9999 & $-1{\times}10^{-4}$ & 0.0032 & 0.0032 & 0.0125 & 1.812 \\[1pt]
 \underline{Oracle MADCovar} & 4.0001 & $+1{\times}10^{-4}$ & 0.0032 & 0.0031 & 0.0122 & 1.812 \\[1pt]
 \underline{Oracle ARMM}     & 3.9999 & $-1{\times}10^{-4}$ & 0.0032 & 0.0032 & 0.0125 & 1.812 \\[1pt]
\bottomrule
\end{tabular}
\end{table}
\begin{table}[ht]
\centering
\caption{%
  Final ATE estimates after $T{=}200$ batches of $n{=}1000$ units, 
  $p{=}100$ covariates, $B{=}2000$, variable costs 
  $c_j \overset{iid}{\sim} \mathrm{Uniform}(0,2)$.
  True ATE $= 4$. Single run, \texttt{set.seed(7)}.
  \underline{DiM} is the baseline benchmark; oracle methods use the
  true 20 signal covariates and are separated by a partial rule.
  Bold marks the best non-benchmark, non-oracle value.
  SE is the HC2 posterior standard error;
  CI Width is $2 \times 1.96 \times \text{SE}$;
  Rel.\ SE $= \hat{\sigma}_{\rm DiM} / \hat{\sigma}_{\rm method}$
  (values ${>}1$ indicate lower estimated variance than DiM).
  All CIs are 95\% (valid at fixed $T$).}
\label{tab:dgp1-performance-single-run}
\small
\setlength{\tabcolsep}{5pt}
\begin{tabular}{@{} l ccccc @{}}
\toprule
\textbf{Method}
  & \textbf{ATE} & \textbf{Bias} & \textbf{SE}
  & \textbf{CI Width} & \textbf{Rel.\ SE} \\
\midrule
 \underline{DiM}      & 4.01222 & $+1{\times}10^{-2}$ & 0.00579 & 0.02270 & 1.0000          \\[1pt]
 ARMM                 & 4.00945 & $+9{\times}10^{-3}$ & 0.00557 & 0.02181 & 1.0404          \\[1pt]
 MADCovar             & 3.99432 & $-6{\times}10^{-3}$ & 0.00550 & 0.02154 & 1.0535          \\[1pt]
 DARTS                & 4.00322 & $+3{\times}10^{-3}$ & 0.00393 & 0.01539 & \textbf{1.4743} \\[1pt]
\cmidrule(l){1-6}
 \underline{Oracle DARTS}    & 4.00319 & $-2{\times}10^{-3}$ & 0.00319 & 0.01249 & 1.8175 \\[1pt]
 \underline{Oracle ARMM}     & 4.00458 & $+5{\times}10^{-3}$ & 0.00318 & 0.01246 & 1.8221 \\[1pt]
 \underline{Oracle MADCovar} & 4.00236 & $+2{\times}10^{-3}$ & 0.00311 & 0.01220 & 1.8598 \\[1pt]
\bottomrule
\end{tabular}
\end{table}
\clearpage
\newpage

\end{document}